\documentclass[letterpaper]{article} 
\usepackage{aaai23}  
\usepackage{times} 
\usepackage{helvet}  
\usepackage{courier}  
\usepackage[hyphens]{url}  
\usepackage{graphicx}
\usepackage{natbib}  
\usepackage{caption}
\usepackage{algorithm}
\usepackage{algorithmic}
\usepackage{amsmath}
\usepackage{amsfonts}

\graphicspath{{./figures/}{../figures/}{./}}

\DeclareMathOperator{\R}{\mathbb{R}}
\DeclareMathOperator{\Mcal}{\mathcal{M}}
\DeclareMathOperator{\Ncal}{\mathcal{N}}
\DeclareMathOperator{\Ucal}{\mathcal{U}}
\DeclareMathOperator{\Vcal}{\mathcal{V}}

\title{Experimental Observations of the Topology of\\Convolutional Neural Network Activations\thanks{Version including technical appendix can be found on ArXiv.}}
\author{
    Emilie Purvine,\textsuperscript{\rm 1}\thanks{Primary; all other authors listed in alphabetical order}
    Davis Brown,\textsuperscript{\rm 1}
    Brett Jefferson,\textsuperscript{\rm 1}
    Cliff Joslyn,\textsuperscript{\rm 1}
    Brenda Praggastis,\textsuperscript{\rm 1}\\
    Archit Rathore,\textsuperscript{\rm 2}
    Madelyn Shapiro,\textsuperscript{\rm 1}
    Bei Wang,\textsuperscript{\rm 2}
    Youjia Zhou\textsuperscript{\rm 2}
}
\affiliations {
    \textsuperscript{\rm 1} Pacific Northwest National Laboratory\\
    \textsuperscript{\rm 2} Scientific Computing and Imaging (SCI) Institute and School of Computing, University of Utah\\
    \{emilie.purvine, davis.brown, brett.jefferson, cliff.joslyn, brenda.praggastis, madelyn.shapiro\}@pnnl.gov, architrathore1@gmail.com, \{beiwang, 
    zhou325\}@sci.utah.edu
}

\begin{document}

\maketitle

\begin{abstract}
Topological data analysis (TDA) is a branch of computational mathematics, bridging algebraic topology and data science, that provides compact, noise-robust representations of complex structures.
Deep neural networks (DNNs) learn millions of parameters associated with a series of transformations defined by the model architecture, resulting in high-dimensional, difficult-to-interpret internal representations of input data.
As DNNs become more ubiquitous across multiple sectors of our society, there is increasing recognition that mathematical methods are needed to aid analysts, researchers, and practitioners in understanding and interpreting how these models' internal representations relate to the final classification.
In this paper, we apply cutting edge techniques from TDA with the goal of gaining insight into the interpretability of convolutional neural networks used for image classification.
We use  two common TDA approaches to explore several methods for modeling hidden-layer activations as high-dimensional point clouds, and provide experimental evidence that these point clouds capture valuable structural information about the model's process. 
First, we demonstrate that a distance metric based on persistent homology can be used to quantify meaningful differences between layers, and we discuss these distances in the broader context of existing representational similarity metrics for neural network interpretability.
Second, we show that a mapper graph can provide semantic insight into how these models organize hierarchical class knowledge at each layer.
These observations demonstrate that TDA is a useful tool to help deep learning practitioners unlock the hidden structures of their models. 
\end{abstract}


\section{Introduction}
Convolutional neural networks (CNNs) are a class of deep learning (DL) models that have been widely used for image classification tasks with great success, but the reasoning behind their decisions is often difficult to determine.
Recent work has established an active field of explainable DL to tackle this problem.
There are tools that highlight areas of the images most influential to the classification \cite{selvaraju2017grad}, or reconstruct idealized input images for each output class \cite{mahendran2015understanding, wei2015understanding}.
There are even tools that try to impose human concepts on the DL model \cite{kim2018interpretability}. 
The complexity and dependencies present within these trained models demand methods in explainable DL that can summarize complex data without losing critical structures, producing features of internal representations that are both stable and persistent with respect to changing inputs and noise, and significant with respect to representing meaningful features of the input data. 

Topological data analysis (TDA) is an emerging field that bridges algebraic topology and computational data science. 
One of the hallmarks of TDA is its ability to provide compact, noise-robust representations of complex structures within data.
These are exactly the kind of representations that are needed in the DL  space where different training runs or noisy input data may result in slightly different hidden activations but in no change in the ultimate classification.
In other well-documented cases, slight changes in input, perhaps unseen to the human eye, result in misclassifications. 
We believe TDA can help us understand these cases as well by recognizing changes in the compact representations of the complex structures of hidden activation layers.

In this paper, we build upon others' recent work in using TDA to understand various aspects of machine learning (ML) and DL models. 
We provide experimental results that show how a topological viewpoint of hidden-layer activations can summarize and compare the complex structures within them and how the conclusions align with our human understanding of the image classification task.
We begin by providing some preliminaries on CNNs and TDA and summarize related work. 
We then show our experiments, which use two tools from TDA: persistent homology and mapper. 
Finally, we conclude with a discussion and our directions for future work.


\section{Preliminaries}

\subsection{Convolutional Neural Networks}
CNNs are a type of deep neural network that respects the spatial information existing in the input data.
They use shared weights to provide translation invariant measures of correlation across an input, which makes them ideal for image classification tasks, where objects requiring identification might be found anywhere in an image.

Mathematically, a trained neural network used for classification is best described as the composition of linear and non-linear \emph{tensor maps} called \emph{layers}, where a \emph{tensor} is a multi-dimensional real-valued array.
The input to a neural network is a tensor, and the output of the network is a probability vector indicating the likelihood the input belongs to each class.
The intermediate outputs from each layer of the composition are called feature maps or \emph{activation tensors}. 
Linear layers use tensor maps that respect element-wise addition and scalar multiplication,
and can be either fully connected or convolutional.

\begin{figure}[t]
    \centering
    \includegraphics[width=0.9\columnwidth]{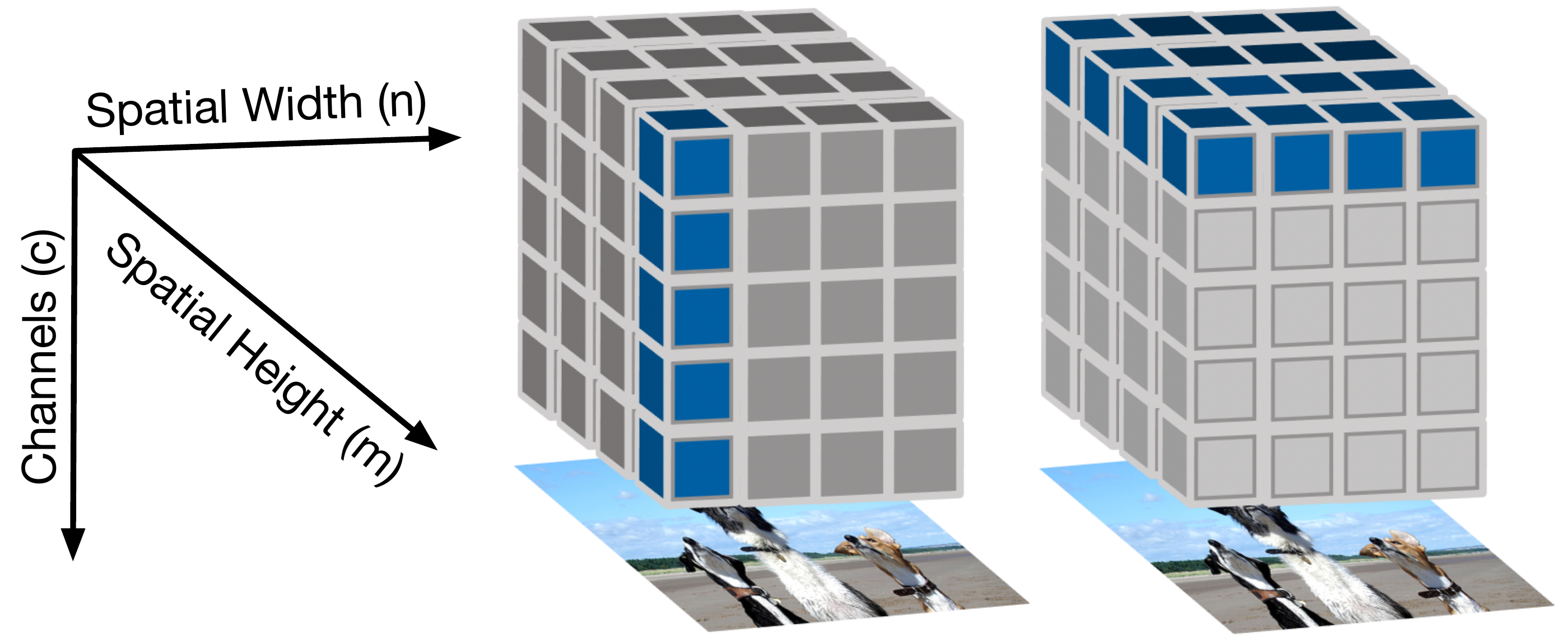}
    \caption{Visualization of spatial activation (middle) and channel activation (right) within an activation tensor.}
    \label{fig:spatial-channel-activations}
\end{figure}

Convolutional layers use cross correlation, also known as a sliding dot product, to map 3D tensors to 3D tensors. 
If the activation tensor from a convolutional layer has dimensions $c \times n \times m$, we say the tensor has
 $c$ \emph{channels} and $nm$ \emph{spatial dimensions}. 
Activation tensors may be sliced into spatial and channel activations, as shown in Figure \ref{fig:spatial-channel-activations}, and then reshaped to obtain vector representations of their values. 

\subsection{Persistent Homology}
One of the two topological tools that we use in our work is persistent homology (PH).  
At a high level, PH is a method for understanding the topological structure of a space that data are sampled from.
We typically have access only to the sample, in the form of a point cloud, and use PH to infer large-scale structures of the unknown underlying space.
Here, we provide a brief overview of PH and point readers to \citet{edelsbrunner2008persistent,ghrist2008barcodes} for more details.

The theoretical basis for persistent homology lies in the concept of \emph{homology} from algebraic topology. 
Given a topological object, e.g., a surface or the geometric realization of a \emph{simplicial complex} (a collection of finite sets, $\Sigma$, such that if $\tau \subset \sigma$ and $\sigma \in \Sigma$ then $\tau \in \Sigma$), its homology is an algebraic representation of its cycles in all dimensions. 
In dimensions 0, 1, and 2, the cycles have simple interpretations as connected components, loops, and bubbles, respectively. 
Higher dimensional interpretations exist but are less intuitive. 

\begin{figure}[t]
    \centering
    \includegraphics[width=1.0\columnwidth]{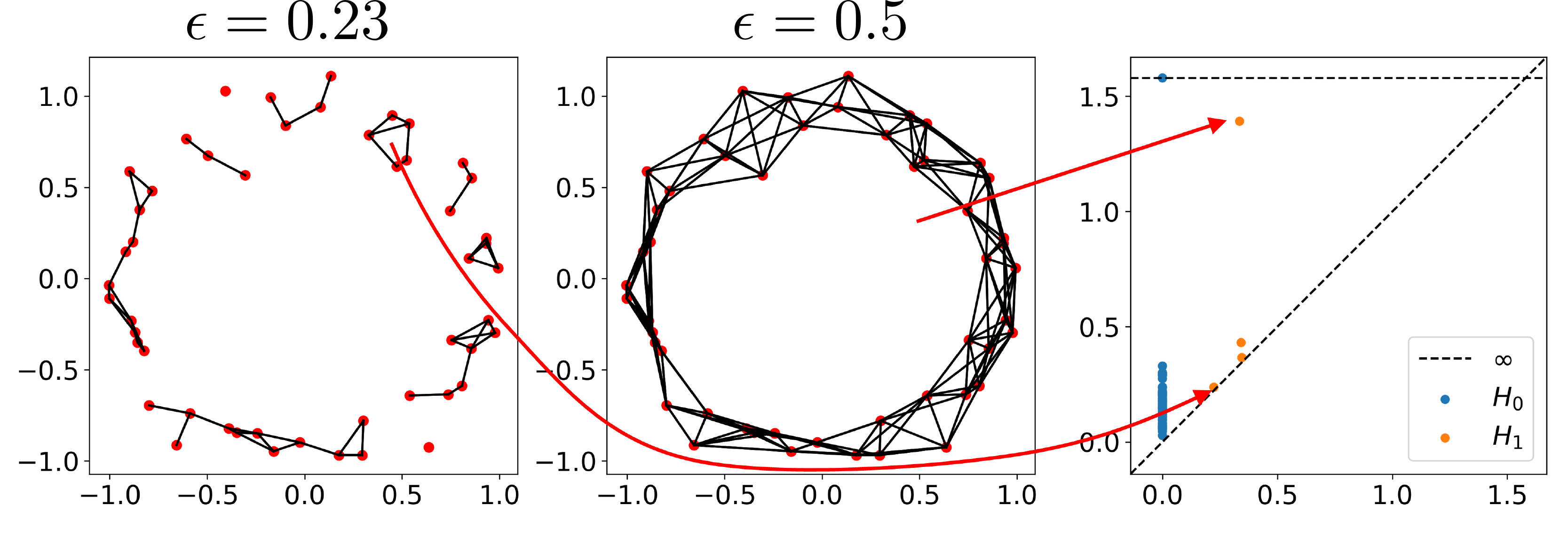}
    \caption{VR complexes at two $\epsilon$ values and the PD of the point cloud. In the PD, orange (resp. blue) points represent 1D (resp. 0D) persistent features. Points on the horizontal dotted line are those that persist through the entire filtration and have no death threshold.}
    \label{fig:VR-complexes}
\end{figure}

Given a single point cloud, $S \subset \R^k$, we can construct a family of associated simplicial complexes on which to compute homology. 
In this paper, we use the Vietoris-Rips (VR) complex given a scale parameter $\epsilon$, $VR(S, \epsilon)$. 
In short, $VR(S, \epsilon)$ is a simplicial complex where each collection of points in $S$ whose pairwise distances are all at most $\epsilon$ is a set in $VR(S, \epsilon)$.
We show examples of two VR complexes (just the 1-skeleton, the pairwise edges) of the same point cloud at two scale parameters in Figure \ref{fig:VR-complexes}.

Finally, we can describe the motivation and concept of PH. 
A single point cloud technically is a simplicial complex, but it is not interesting homologically.
Whereas constructing a VR complex at a single scale parameter does provide an interesting topological object, it does not capture the multiscale phenomena of the data.
PH is a method that considers all VR scale parameters together to identify at which $\epsilon$ a cycle is first seen (is ``born'') and at which $\epsilon'$ the cycle is fully triangulated (``dies''). 
This set of birth and death values for a sequence of simplicial complexes of a given point cloud provides a topological fingerprint for a point cloud often summarized in a \emph{persistence diagram} (PD) as a set of $(b, d)$ coordinates.
Figure \ref{fig:VR-complexes} also shows the point cloud's PD from the full sequence of $\epsilon$ thresholds.

PDs form a metric space under a variety of distance metrics.
In this paper, we will use \emph{sliced Wasserstein (SW) distance} introduced by \citet{pmlr-v70-carriere17a}.
Given two PDs, the SW distance is computed by integrating the Wasserstein distances for all projections of the PD onto lines through the origin at different angles.

\subsection{Mapper}
The mapper algorithm was first introduced by \citet{singh2007topological}. It is rooted in the idea of ``partial clustering of the data guided by a set of functions defined on the data''~\shortcite{singh2007topological}.
On a high level, the mapper graph captures the global structure of the data.

Let $S \subset \R^k$ be a high-dimensional point cloud. A \emph{cover} of $S$ is a set of open sets in $\R^k$,  $\Ucal = \{U_i\}$ such that $S \subset \cup_{i} U_i$. 
In the classic mapper construction, obtaining a cover of $S$ is guided by a set of scalar functions defined on $S$, referred to as \emph{filter functions}. For simplicity, we describe the mapper construction using a single filter function $f: S \to \R$. Given a cover $\Vcal = \{V_\ell\}$ of $f(S) \subset \R$ where $f(S) \subseteq \cup_{\ell} V_\ell$, we can obtain a cover $\Ucal$ of $S$ by considering as cover elements the clusters (for a choice of clustering algorithm) induced by $f^{-1}(V_\ell)$ for each $V_\ell$. 

Then, the 1D \emph{nerve} of any cover $\Ucal$ is a graph and is denoted as $\Ncal_1(\Ucal)$. Each node $i$ in $\Ncal_1(\Ucal)$ represents a cover element $U_i$, and there is an edge between nodes $i$ and $j$ if $U_i \cap U_j$ is non-empty. 
If $\Ucal$ is constructed as above, from a clustering of preimages of a filter function $f$, then its 1D nerve, denoted as $\Mcal = \Mcal(S, f):= \Ncal_1(\Ucal)$, is the \emph{mapper graph} of $(S, f)$.

\begin{figure}[t]
 \centering
 \includegraphics[width=.9\columnwidth]{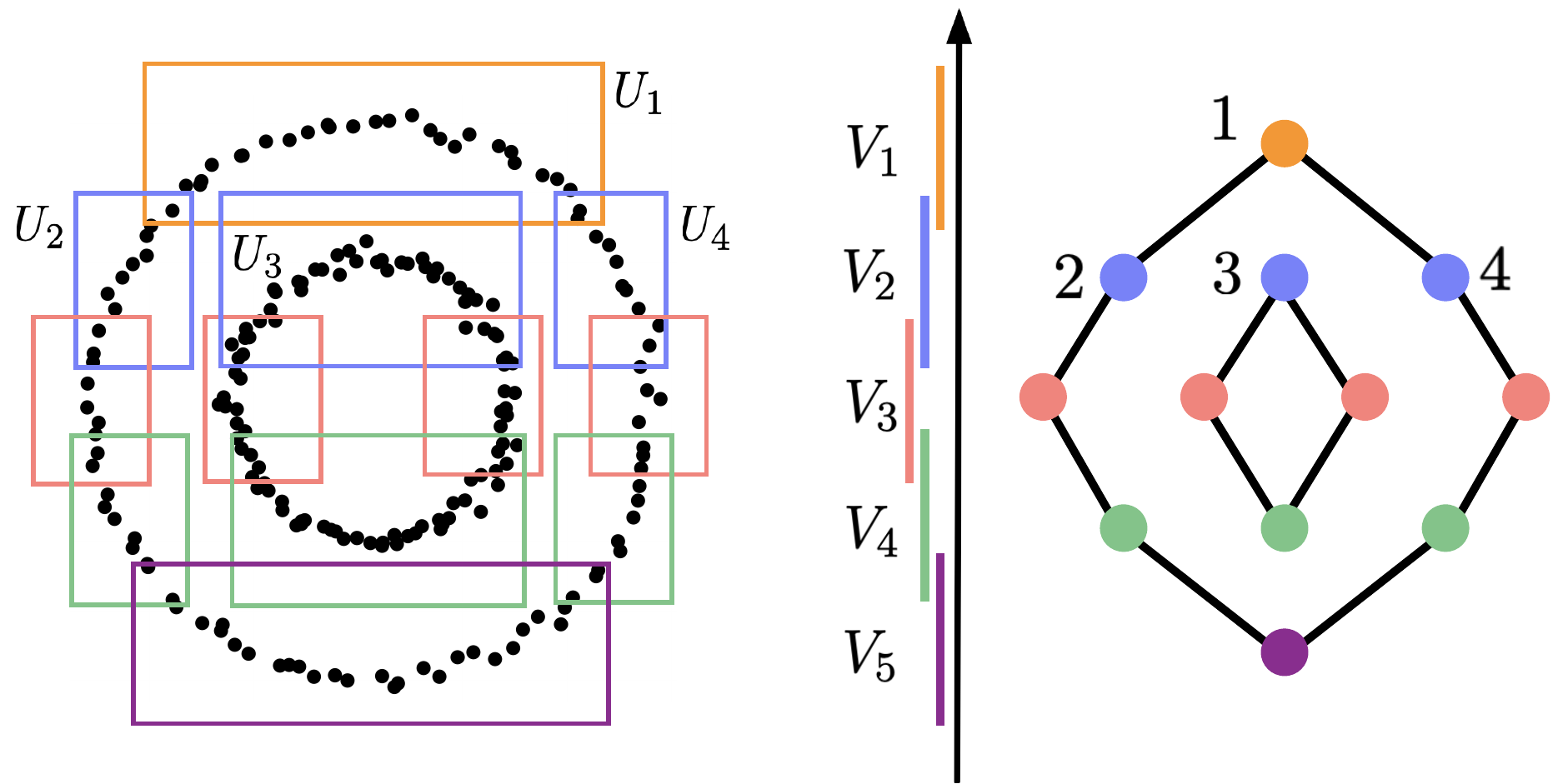}
 \caption{A mapper graph of a point cloud containing two nested circles.}
 \label{fig:two-circles}
\end{figure}

Consider the point cloud in Figure~\ref{fig:two-circles} as an example containing two nested circles. It is equipped with a height function $f: S \to \R$. A cover $\Vcal = \{V_1, \cdots, V_5\}$ of $f(S)$ is formed by five intervals (see Figure~\ref{fig:two-circles} middle). For each $\ell$ ($1 \leq \ell \leq 5$), $f^{-1}(V_\ell)$ induces a number of clusters that are subsets of $S$. Such clusters form the elements of a cover $\Ucal$ of $S$. As shown in Figure~\ref{fig:two-circles} (left), the cover elements of $\Ucal$ are contained within the 12 rectangles on the plane. The mapper graph of $S$ is shown in Figure~\ref{fig:two-circles}c. For instance, cover $f^{-1}(V_1)$ induces a single cover element $U_1$ of $S$, and it becomes node 1 in the mapper graph of $S$. $f^{-1}(V_2)$ induces 3 cover elements $U_2$, $U_3$ and $U_4$, which become nodes 2, 3 and 4. Since $U_1 \cap U_2 \neq \emptyset$, an edge exists between node 1 and node 2. 
The two circular structures in Figure~\ref{fig:two-circles} (left) are captured by the mapper graph in Figure~\ref{fig:two-circles} (right).

\subsection{Related Work}
The value of TDA to organize, understand, and interpret various aspects of ML and DL models has been recognized in several current research directions. 
Much of this research has focused on model parameters, structure, and weights.
\citet{guss2018characterizing} examine model architecture selection by defining the ``topological capacity'' of networks, or the ability for the network to capture the true topological complexity of the data.
They explore the learnability of model architectures in the face of increasing topological complexity of data. 
\citet{gabrielsson2019exposition} build the mapper graph of a point cloud of learned weights from convolutional layers within a simple CNN and find that the weights of different CNN model architectures trained on the same data set have topological similarities. 
``Neural persistence'', developed by \citet{rieck2019neural}, is a topological measure of complexity of a fully connected deep neural network that depends on learned weights and network connectivity. 
They find networks that use best practices such as dropout and batch normalization have statistically higher neural persistence, and define a stopping criterion to speedup the training of such a network. 

Other studies, like that of \citet{wheeler2021activation} use TDA to study activation tensors of simple multi-layer perceptron networks to discover how the topological complexity, as measured by a property of persistence landscapes, changes through the layers. \citet{gebhart2019characterizing,lacombe2021topological} investigate the topology of neural networks via ``activation graphs,'' which model the natural graphical structure of the network.
Finally, most closely related to our work is that of \citet{rathore2021topoact}, which describes TopoAct, a visual platform to explore the organizational principle behind neuron activations. 
TopoAct displays the mapper graph of activation vectors for a single layer at a time in a CNN to show how the model organizes its knowledge via the branching structures. 
The authors consider a point cloud formed by randomly sampling a single spatial activation in a given layer for each image in a corpus.
We extend this work by using a larger and more data-driven sample of spatial activations to build our mapper graphs, quantifying the intuition of ``pure'' and ``mixed'' mapper nodes, considering the effect of noisy input on the resulting graph, and showing how our results generalize to multiple common model architectures.

\section{Point Cloud Summaries of Activations}
\label{sec:point_cloud_activation}

Following the approach of \citet{rathore2021topoact}, we model each convolutional layer of a CNN as an $Np\times c$ point cloud by sampling $p$ spatial activation vectors from the $c\times n\times m$ activation tensors produced by $N$ images in a dataset. This gives us a collection of point clouds that can be used to study the evolution of the activation space (i.e., the space of spatial activations), as the complexity of features learned by each layer increases as we move deeper into the model \cite{zhou2015object,olah2020zoom}. We introduce several data-driven sampling methods with the goal of improving upon the quality of the sampled point cloud representation.

\subsubsection{Random and full activations.} 
In our mapper experiments, for a fixed layer, we construct a high-dimensional point cloud by \emph{randomly sampling} a single ($p=1$) spatial activation from each input image, as in \citet{rathore2021topoact}. 
We additionally experiment with \emph{full activation sampling} ($p=nm$) by including all spatial activations of a given layer for each image in the point cloud construction.

\subsubsection{Top $l^2$-norm activations.}
In our PH experiments, for a fixed layer we construct a point cloud with \emph{top $l^2$-norm sampling} ($p=1$) by selecting the spatial activation with the strongest $l^2$-norm from each image.

\subsubsection{Foreground and background activations.}
\begin{figure}[t]
    \centering
    \includegraphics[width=.9\columnwidth]{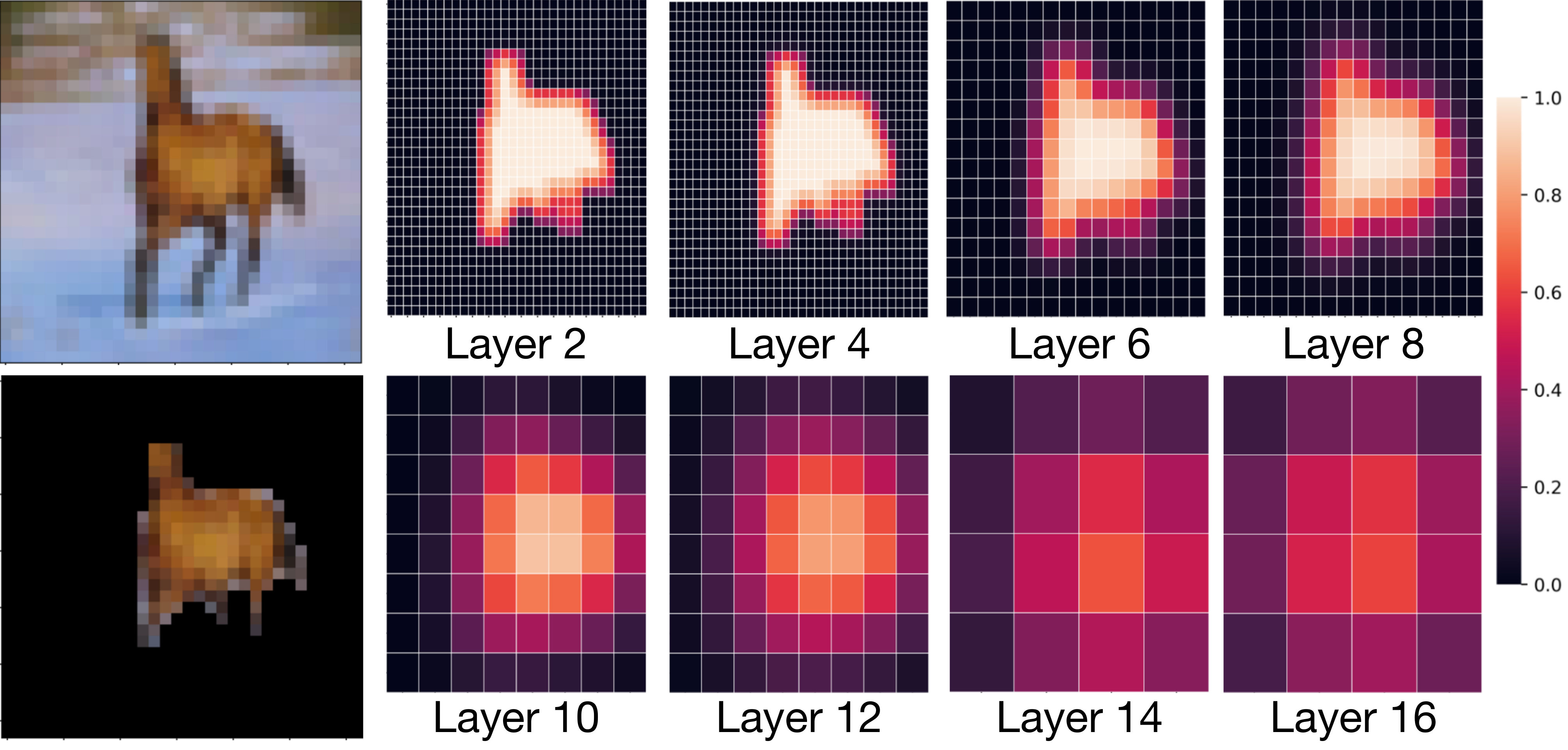}
    \caption{Spatial positions whose effective receptive field contains primarily foreground pixels are highly weighted in foreground sampling.}
    \label{fig:foreground-detection}
\end{figure}

For a fixed convolutional layer, each spatial position in the activation tensor can be traced back to its \emph{effective receptive field}, which is the region of the input image that the network has ``seen'' via contributions from previous layers. Naturally, each spatial activation corresponds to the subset of the foreground and background pixels in its effective receptive field. To investigate how foreground and background information of an input image manifests in the activation space, we first use {\tt cv2.grabCut} from the OpenCV library \cite{opencv_library} to perform image segmentation and identify the foreground and background pixels in the images. We then assign a weight to each spatial activation according to the number of foreground or background pixels in its effective receptive field, as illustrated in Figure \ref{fig:foreground-detection}. The spatial activations with the greatest weight are selected to represent each image in the point cloud construction, referred to as \emph{foreground} or \emph{background sampling}. In our mapper experiments, we study the ``top $p$'' foreground and background activations for $p=1$ and $p=5$. 

\subsection{Reproducibility Details}
The following two sections outline our experiments using PH and mapper graphs to study the standard benchmark dataset CIFAR-10~\cite{cifar10} on a ResNet-18 architecture~\cite{resnet}. We perform standard preprocessing to normalize the images by the mean and variance from the full training set. Code for the models and additional details regarding the dataset, as well as the parameters and computing infrastructure specific to each set of experiments, are provided in the arXiv technical appendix.

\section{Experiments with PH}

Using the top $l^2$-norm sampling method, we construct point cloud summaries of activations from the CIFAR-10 dataset on a ResNet-18 model to study the PH of the activation space.
The SW distance between PDs of these point cloud summaries --- which we will refer to from now on as the \emph{SW distance between layers} --- proves to be an interesting topological metric for capturing similarity between layers; it exhibits some of the fundamental qualities of strong representation similarity metrics for neural networks but fails to be sensitive to others \cite{ding2021grounding}.

\subsection{Relationships Between Layers}
\begin{figure}[t]
    \centering
    \includegraphics[width=1.0\columnwidth]{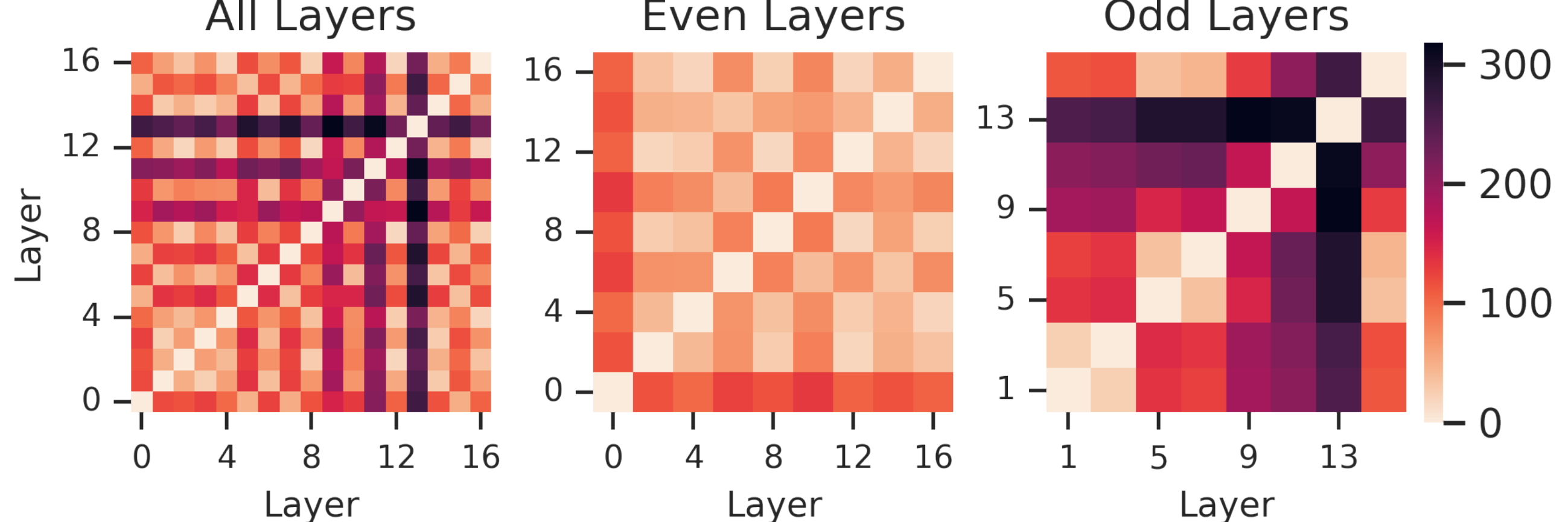}
    \caption{SW distances between convolutional layers of ResNet-18; results averaged over 10 random batches of 1000 CIFAR-10 test set images ($\text{CV}<0.17$).}
    \label{fig:sw-heatmap-rn18}
\end{figure}

In Figure \ref{fig:sw-heatmap-rn18}, we observe a grid-like pattern in the SW distances between layers of ResNet-18 similar to the results found in \citet{kornblith2019similarity}, which the authors attribute to the residual architecture. This observation supports our belief that meaningful qualities of the model and its architecture can be uncovered by studying the topology of the activation space with PH.

\subsection{Representation Similarity Metrics \& Intuitive Tests}
Metrics such as canonical correlation analysis (CCA)~\cite{morcos2018insights,raghu2017svcca}, centered kernel alignment (CKA)~\cite{kornblith2019similarity}, and orthogonal Procrustes distance~\cite{ding2021grounding} provide dissimilarity measures that can be used to compare layers of neural networks. Recent work has demonstrated the value of topological approaches to representation similarity such as Representation Topology Divergence \cite{barranikov2022representation}. These methods operate on an $N\times cnm$ matrix representation of a convolutional layer, where the $c\times n\times m$ activation tensors produced by each of the $N$ inputs from the dataset are normalized and unfolded into vectors in $\R^{cnm}$. Here we note this as a key difference from our $N\times c$ point cloud representation obtained through top $l^2$-norm sampling but leave a more thorough comparison to future work.

We apply the intuitive specificity and sensitivity tests outlined by \citet{ding2021grounding} to probe the utility of the SW distance between layers as a representation similarity metric for neural networks. In comparison to the intuitive test results shown for CCA, CKA, and orthogonal Procrustes distance from \citet{ding2021grounding}, this metric exhibits some non-standard behavior, for which we provide some speculative explanations but further work is needed to fully understand such a metric.

\subsubsection{Specificity.}
\begin{figure}[t]
    \centering
    \includegraphics[width=1.0\columnwidth]{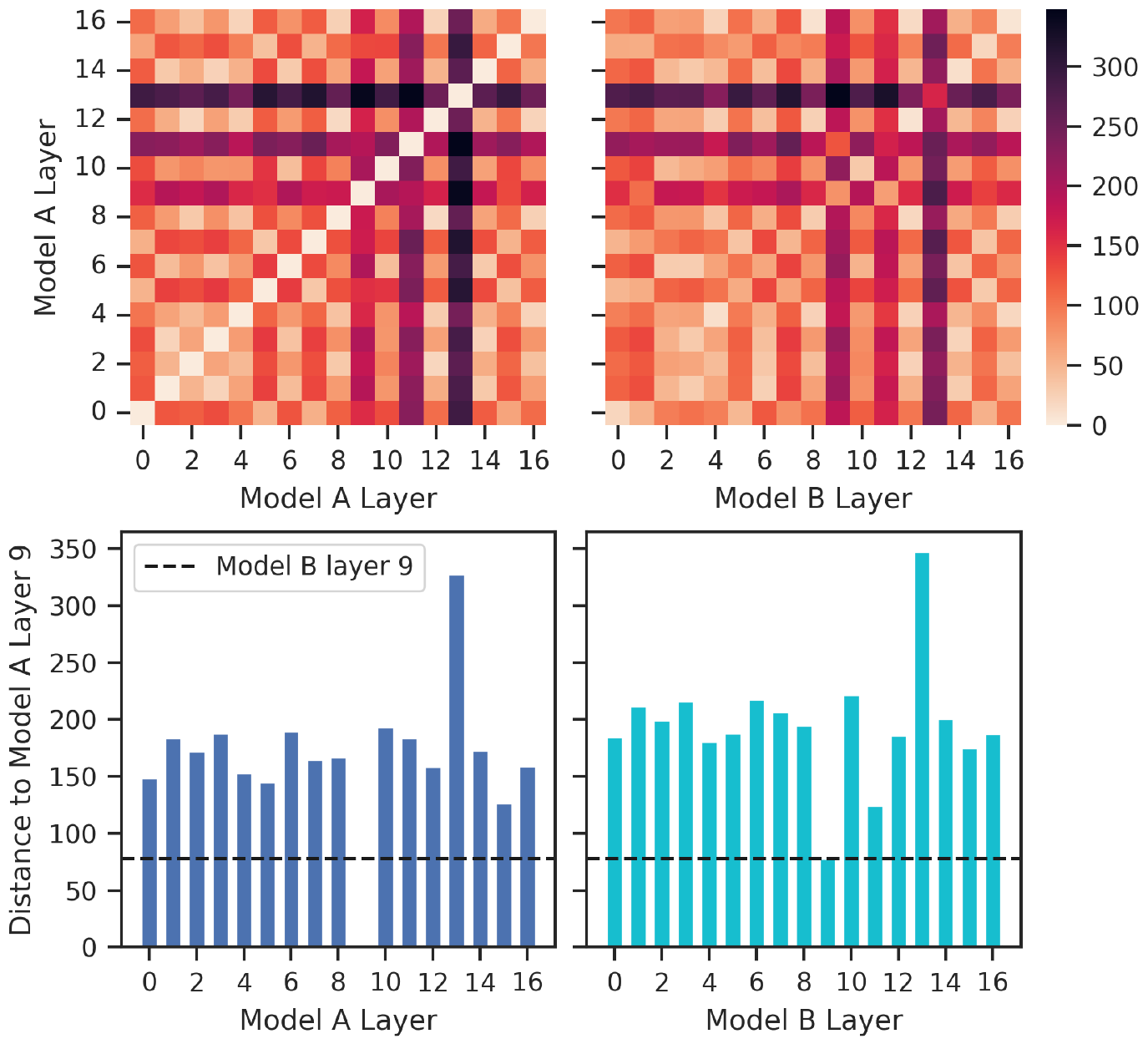}
    \caption{Intuitive specificity test of SW distance between convolutional layers of two ResNet-18 models initialized with different random seeds, for 1000 CIFAR-10 test set images.}
    \label{fig:sw-seed-spec-rn18}
\end{figure}

To measure the impact of model initialization seed on the SW distance between layers, we trained 100 ResNet-18 models with different initialization seeds on CIFAR-10, and constructed top $l^2$-norm point cloud representations of the layers of each model from $N=1000$ test set images. 
Figure \ref{fig:sw-seed-spec-rn18} shows SW distances for two of the models ``A'' and ``B'', comparing pairs of layers in Model A (left) as well as pairs of layers between Model A and Model B (right). We find that variation in model seed has almost no impact on the SW distances, as shown by the near-identical heatmaps and highlighted for layer 9 (bottom row). The internal and cross-model SW distances relative to Model A layer 9 are highly correlated, with $\rho\approx 0.907$ computed by averaging correlation with fixed Model A over the 99 remaining randomly initialized models as Model B. Averaging internal and cross-model correlation relative to each layer of Model A, we find $\rho\approx 0.910$. We conclude that SW distance between layers is highly specific and robust to variation in initialization seed.

\subsubsection{Sensitivity.}
\begin{figure}[t]
    \centering
    \includegraphics[width=1.0\columnwidth]{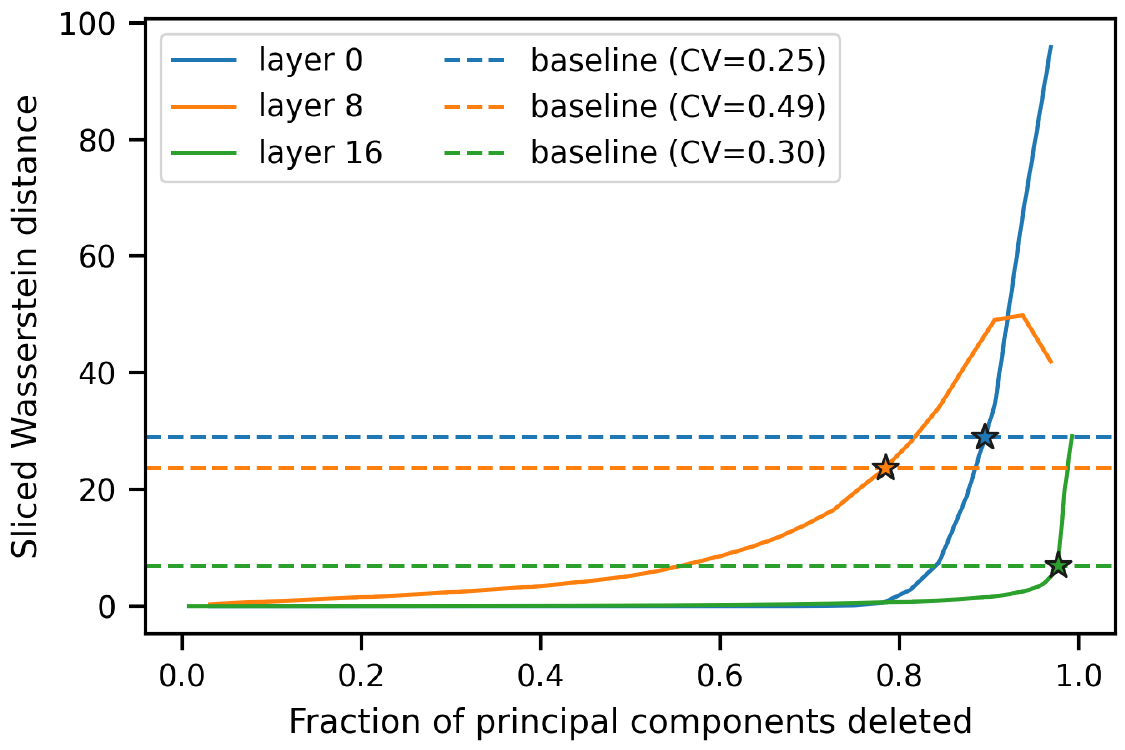}
    \caption{Intuitive sensitivity test of SW distance for the first (0), middle (8), and last (16) convolutional layers of ResNet-18, for 1000 CIFAR-10 test set images.}
    \label{fig:sw-pcasens-rn18}
\end{figure}

A representation similarity metric should be robust to noise without losing sensitivity to significant alterations. We apply the intuitive sensitivity test of \citet{ding2021grounding} by taking the SW distance between each layer and its low-rank approximations as we delete principal components from the $N\times c$ point cloud. The SW distance to the corresponding layer in another model is averaged over the remaining 99 randomly initialized models to compute a baseline SW distance for each layer. This baseline defines a threshold of \emph{detectable} SW distance, above which distance cannot be solely attributed to different initialization. In Figure \ref{fig:sw-pcasens-rn18}, we see the sensitivity of this metric is heavily dependent on layer depth. 

\section{Experiments with Mapper Graphs}
\label{sec:experimens-mapper}

\begin{figure*}[t]
    \centering
    \includegraphics[width=.95\textwidth]{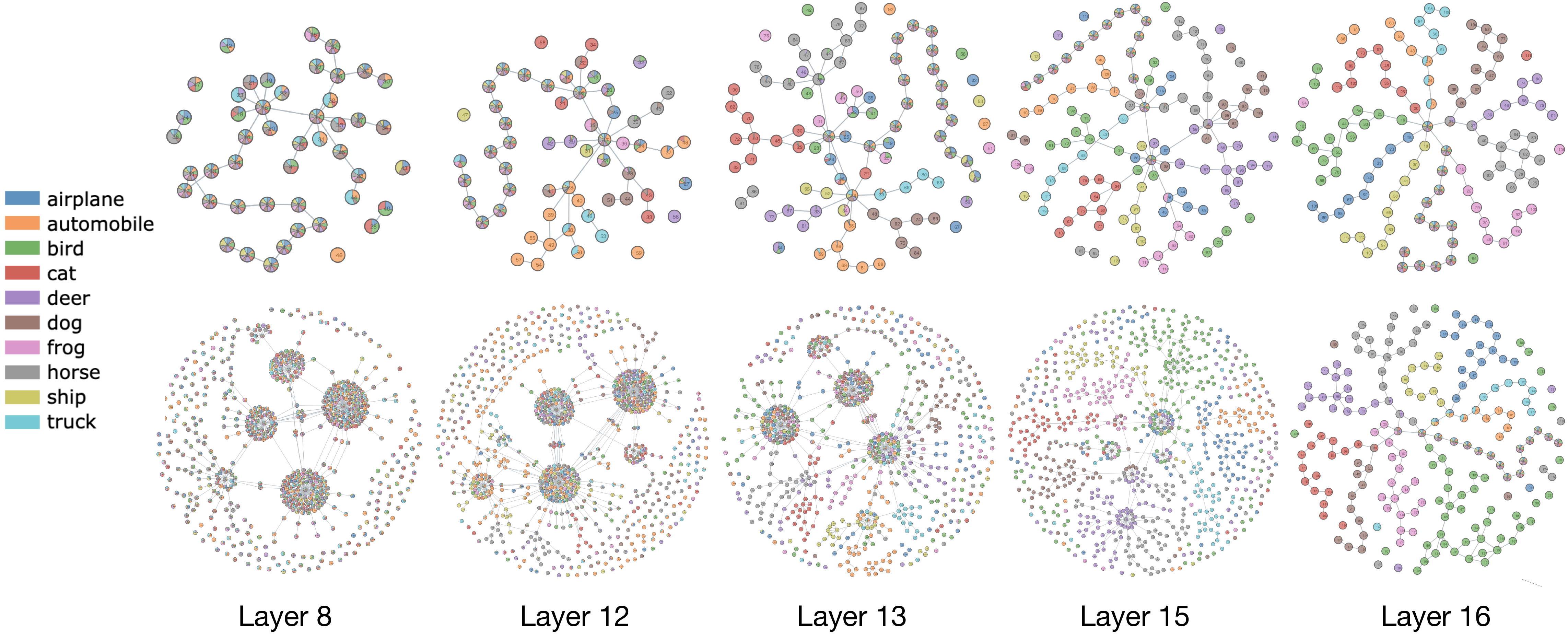}
    \caption{Mapper graphs from random (top) and full (bottom) activations from ResNet-18 using the CIFAR-10 dataset.}
    \label{fig:mapper-random-full}
\end{figure*}

In this section, we explore how the topology of the activation space changes across  layers by constructing mapper graphs from spatial activations from $N=50\text{k}$ CIFAR-10 training images on a ResNet-18 model.
The mapper graph filter function is the $l^2$-norm of each spatial activation. 
We employ and extend \emph{MapperInteractive}~\cite{zhou2021mapper}, an open-source web-based toolbox for analyzing and visualizing high-dimensional point cloud data via its mapper graph. 
Because of the visual nature of mapper graphs, our experiments will largely be evaluated by exploring and comparing the \emph{qualitative} properties of the visualizations rather than quantitative comparisons of structures.
The exception will be our purity measures, introduced in a later subsection.

\subsection{Random and Full Activations}
In Figure~\ref{fig:mapper-random-full}, we compare the mapper graphs generated from a point cloud of random activations ($50\text{k}\times c$) against those generated from the full activations ($50\text{k}\cdot nm\times c$) across different convolutional layers, where $c$ is the number of dimensions of each activation, and $nm$ is the total number of spatial activation vectors per image. 
The glyph for each node of the mapper graph is a pie chart showing the composition of class labels in that node.
It can be seen that at layer 16, the mapper graphs of the random and full activations clearly capture the separation among class labels; there is a central region in the graph where nodes with mixed labels (with lower $l^2$-norm) separate out into branches with single labels (with higher $l^2$-norm). 
As we move toward earlier layers, the ability of the mapper graphs to show class separation gradually deteriorates. In addition, both random and full activations show similar bifurcation patterns, indicating robustness with respect to the sampled activations. 

\subsection{Foreground and Background Activations}
\begin{figure*}[t]
    \centering
    \includegraphics[width=.95\textwidth]{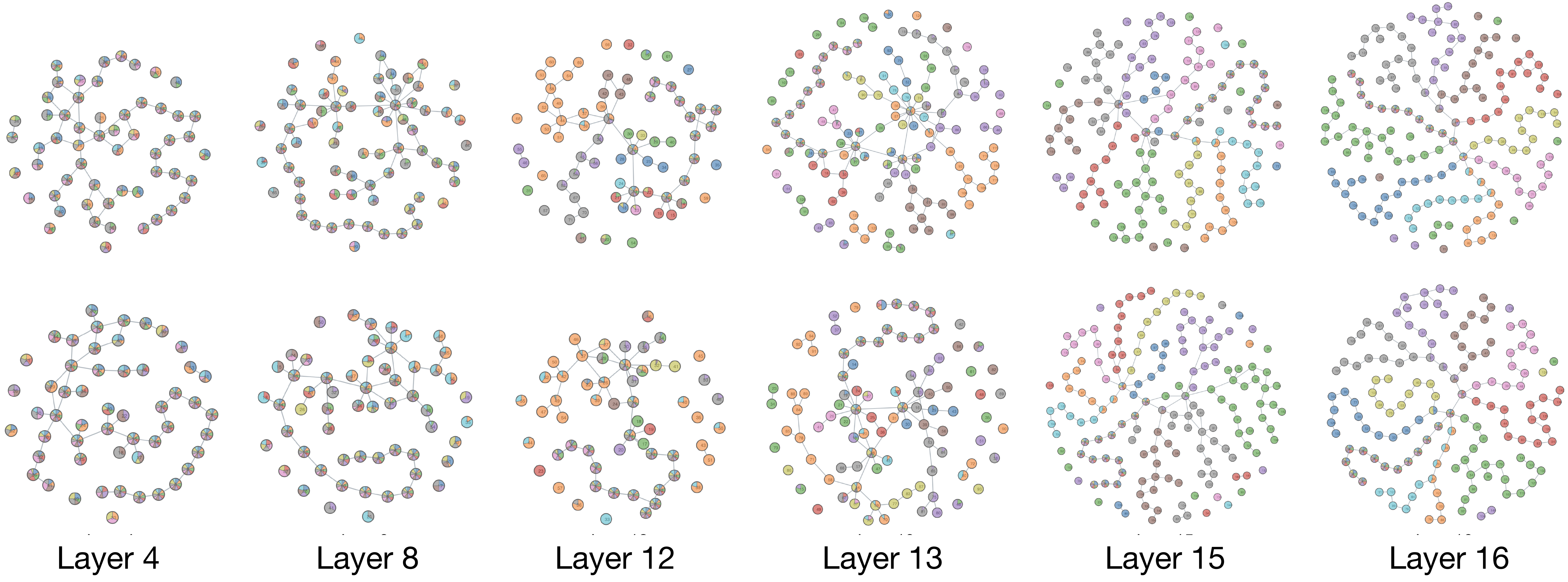}
    \caption{Mapper graphs generated from the foreground (top) and background (bottom) activations with the largest weights.}
    \label{fig:mapper-fg-bg}
\end{figure*}

Next, we study whether branching structures emerge at earlier layers if we use top foreground or background activations. Figure~\ref{fig:mapper-fg-bg} shows the evolution of mapper graphs using the foreground and background activations across layers. We observe that the mapper graph of foreground activations at layer 15 already shows notable class bifurcations. Such early separations are less obvious for random and full activations. The mapper graphs of background activations also show clear class separations at layer 15 and 16, indicating that background pixels likely play an important role in class separation as well. 
Mapper graphs for the top 5 foreground and background activations are provided, along with similar observations in the technical appendix.

\subsection{Activations with Gaussian Noise}

\begin{figure}[t]
    \centering
    \includegraphics[width=0.9\columnwidth]{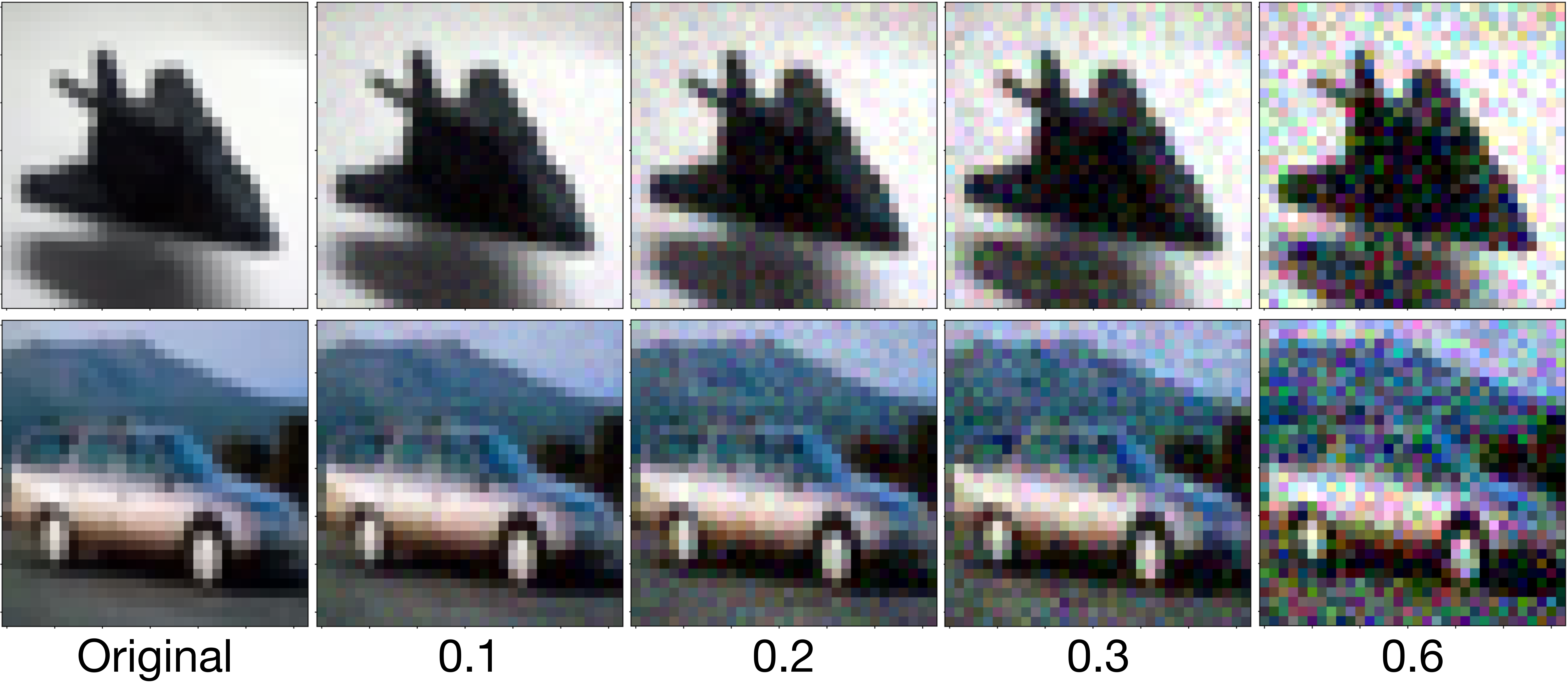}
    \caption{Examples of CIFAR-10 images with perturbations. Column 1 contains the original, and columns 2-4 contain images perturbed with different standard deviations.}
    \label{fig:perturbed-imgs}
\end{figure}

\begin{figure*}[t]
    \centering
    \includegraphics[width=0.95\textwidth]{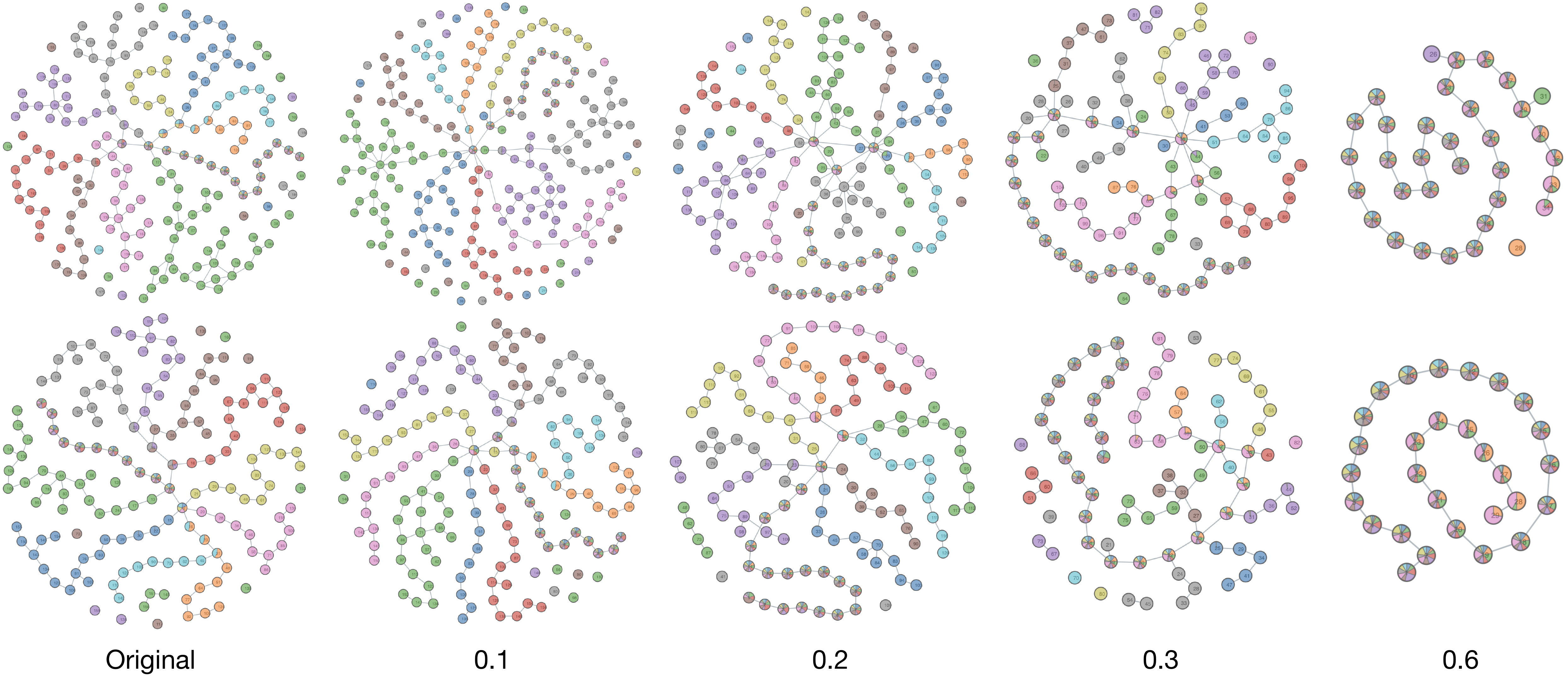}
    \caption{Perturbed mapper graphs generated from the full activations (top) and the foreground activations (bottom) at the last convolutional layer.}
    \label{fig:mapper-noises}
\end{figure*}

To explore the stability of mapper graphs to noise in the input data, we injected pixel-wise Gaussian noise to all 50k images with different standard deviations ($\sigma$). Examples of how the images change as the standard deviation increases are shown in Figure~\ref{fig:perturbed-imgs}, and the corresponding mapper graphs at layer 16 are shown in Figure~\ref{fig:mapper-noises}. It can be seen that the mapper graphs are stable for small perturbations ($\sigma = 0.1$). As $\sigma$ increases, mapper graphs illustrate that the model’s ability to differentiate different classes decreases. This observation aligns with the intuition that increasing the noise level will decrease prediction accuracy.

\subsection{Mapper Graph Purity Measures}
For an image classification task, each point (i.e., a spatial activation) $x \in S$ is assigned a class label (inherited from the class label of its corresponding input image). We introduce three quantitative measures to quantify how well a mapper graph of the activation space separates the points from different classes.  

\subsubsection{Node-wise purity.}
Given a mapper graph $\Mcal$, the node-wise purity of a node $i$ is defined as 
$\alpha_i = \frac{1}{c_i},$ 
where $c_i$ is the number of class labels in node $i$: the more classes in node $i$, the less pure node $i$ is. Figure~\ref{fig:mapper-purity} (bottom) shows the node-wise purity of mapper graphs for foreground (top 1 and 5), random, and full activations at a variety of layers (aligning with the layers seen in Figures \ref{fig:mapper-random-full} and \ref{fig:mapper-fg-bg}). We observe that node-wise purity is larger in deeper layers, indicating that the underlying model gets better at separating the classes the deeper we go.
However, the type of sampling seems not to influence the purity as much. Top 5 foreground sampling tends  to have slightly higher purity, whereas random sampling has lower purity.

\subsubsection{Point-wise purity.}
For a point $x \in S$, the point-wise purity is defined as
$$\beta_x = \frac{\sum_{i=1}^{n_x} \alpha_i}{n_x},$$
where $n_x$ is the number of nodes containing point $x$. It is the average node-wise purity of all nodes containing $x$.

\subsubsection{Class-wise purity.}
\begin{figure}[!ht]
    \centering
    \includegraphics[width=.9\columnwidth]{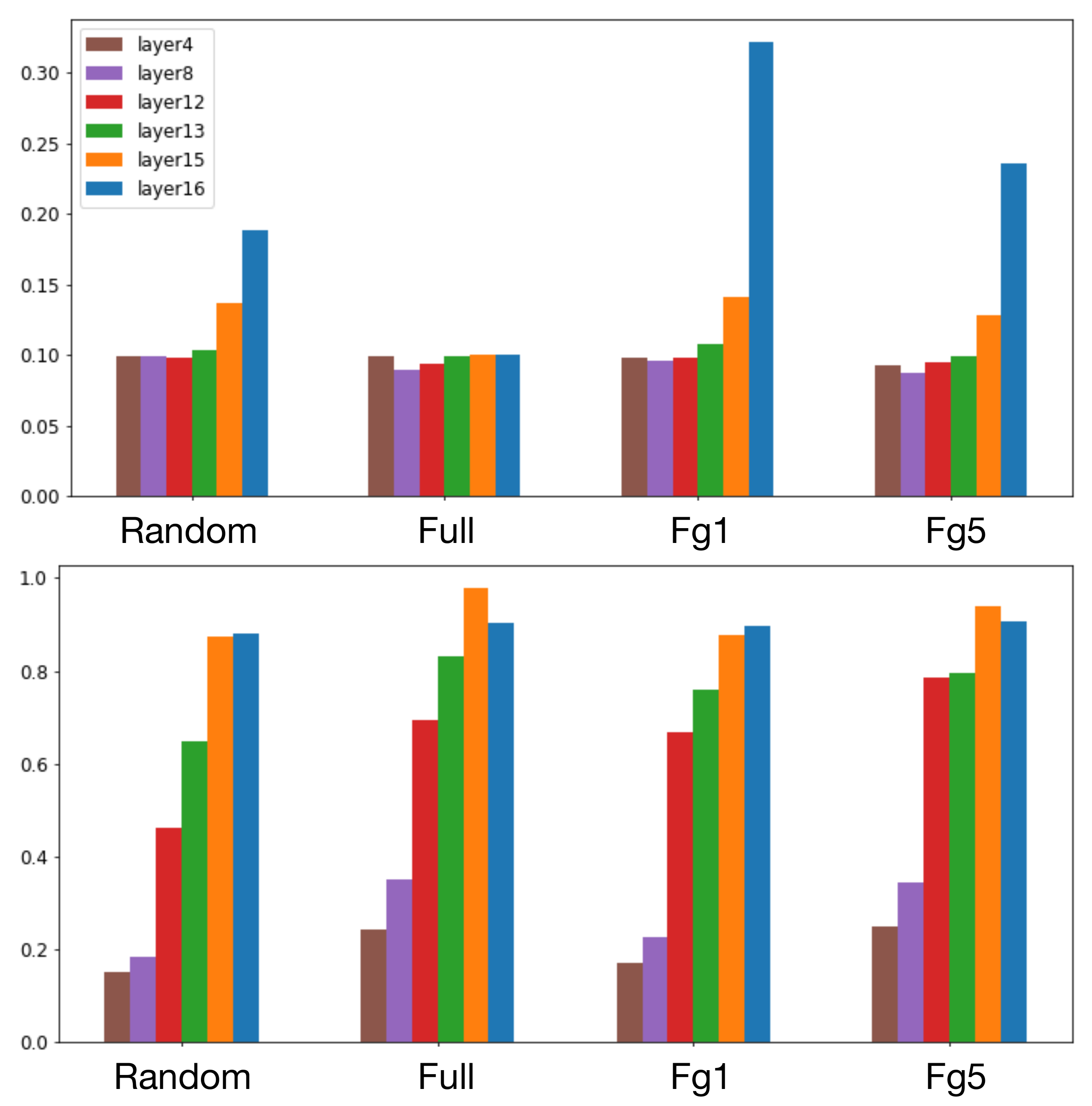}
    \caption{Top: class-wise purity of the deer class for random, full activations, and foreground (top 1 and 5) at a variety of layers; bottom: node-wise purity for random, full activations, and foreground (top 1 and 5) at a variety of layers, and the legend is the same as that of the top plot.}
    \label{fig:mapper-purity}
\end{figure}

For a class $k$, the class-wise purity is defined as 
$$\gamma_k = \frac{\sum_{i=1}^{N_c} \beta_i}{N_c},$$
where $N_c$ is the number of points in class $k$.
It is the average value of point-wise purity for all points in class $k$.
Figure~\ref{fig:mapper-purity} (top) shows the class-wise purity of the deer class for foreground (top 1 and 5), random, and full activations at the same set of layers as node-wise purity. As was the case for the node-wise purity, we observe a general trend of increased class-wise purity of mapper graphs in deeper layers of the neural network.

\subsection{Generalization of Mapper Experiments to Additional Models}

\begin{figure*}[t]
    \centering
    \includegraphics[width=.95\textwidth]{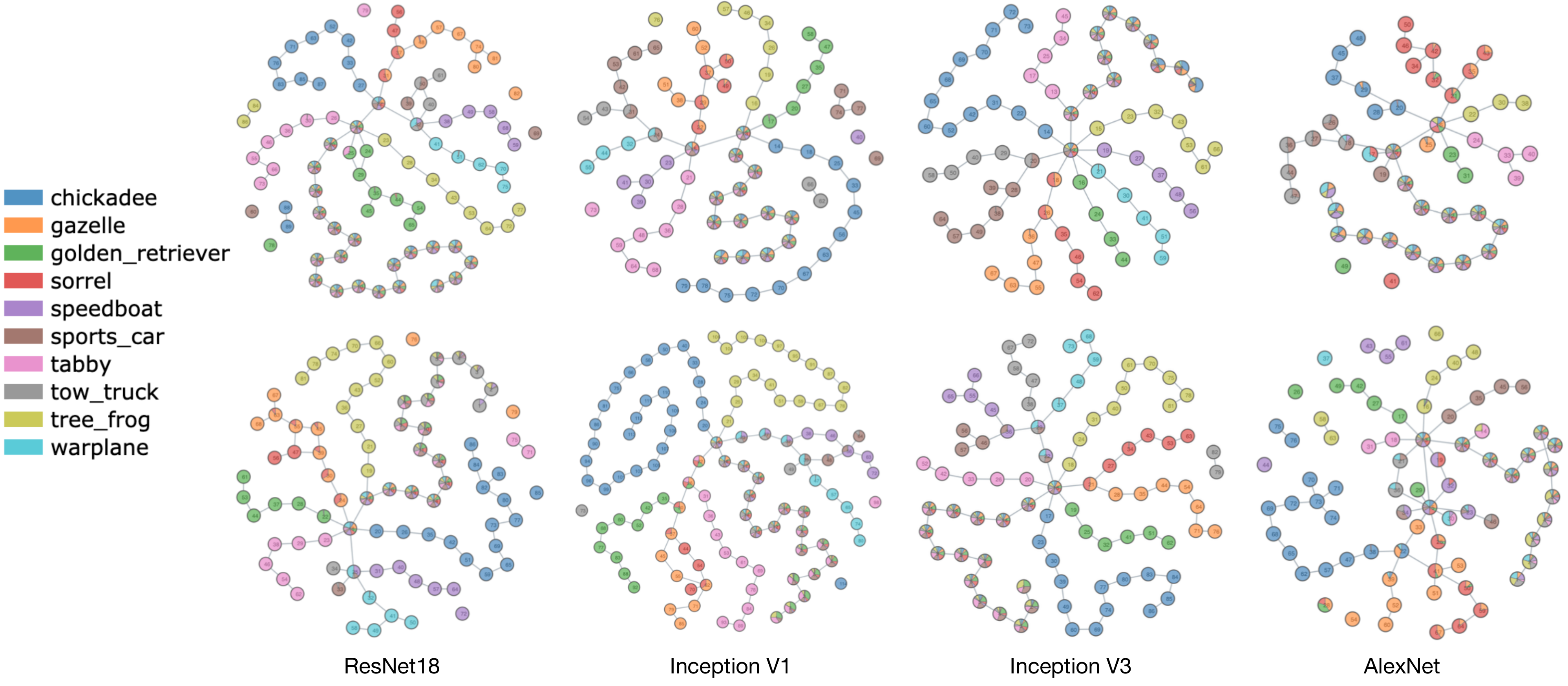}
    \caption{Mapper graphs of random (top) and foreground (bottom) activations for models trained on the ImageNet dataset.}
    \label{fig:mapper-graphs-additional}
\end{figure*}

In order to show that our mapper graph observations are not dependent on the ResNet-18 architecture or CIFAR-10 data set we also perform these experiments using a different model-data pair.
To compare with the prior experiments which use the lower resolution CIFAR-10 data set, the experiments in this section use a subset of 10 classes from the ImageNet dataset \cite{deng2009imagenet}, as shown in the legend of Figure \ref{fig:mapper-graphs-additional}.
There are 1300 images per class, resulting in a set of $N=13\text{k}$ images.
The images have varying resolutions with an average resolution of $469\times 378$.
The data is pre-processed by first resizing each image to 256 pixels and center cropping to a patch of size 2$24\times 224$, followed by a normalization with mean and variance of the original ImageNet training set images.
For foreground extraction, we apply a different strategy than previously used since \texttt{cv2.grabCut} does not work as well with the ImageNet dataset due to the large amount of high frequency details in the image backgrounds.
Instead we use a pre-trained DeepLabV3 semantic segmentation model \cite{chen2017rethinking} to obtain the foreground mask which is then applied to the images to get the foreground pixels.
 
The models that we use for the generalization experiments include ResNet-18, Inception\_v1~\cite{szegedy2015going}, Inception\_v3~\cite{szegedy2016rethinking} and AlexNet~\cite{krizhevsky2012imagenet}. The number of parameters of each model is 11.6M, 6.6M, 27.2M and 61.1M respectively.

Figure~\ref{fig:mapper-graphs-additional} shows the resulting mapper graphs generated from the last layer of each model. Through these experiments, we demonstrate that the structures and insights we observe on ResNet-18 applied to CIFAR-10 are applicable to a wide range of other image recognition models as well.

\section{Discussion and Future Work}
Our experiments using PH and mapper to study activation tensors of CNNs add to the growing body of literature to suggest that TDA provides useful summaries of DL models and hidden representations. 
The ability of mapper graphs to summarize point clouds from activation tensors and identify branching structures was previously shown in \cite{rathore2021topoact}. 
In our paper, we go beyond the random activations of that prior work to build mapper graphs of foreground, background, and full activation point clouds.
These mapper graphs exhibit branching structures at earlier layers and show robustness with respect to  image noise.
Our new purity measures further quantify the observation that mapper graphs' branching structures align with class separations, and improve as we go deeper into the layers.
Moreover, we also show that the mapper graph branching structures are present not just in ResNet-18 applied to CIFAR-10 but also to ImageNet studied using ResNet-18, InceptionV1 and V3, and AlexNet.

Although the mapper graphs we study come from a single trained model, our PH experiments show that the topological structures of the point clouds from which the mapper graphs are built are independent of the training run.
Work has yet to be done to characterize those topological structures for CNNs beyond mapper graphs, but the fact that the distances are training-invariant indicates that such structures are indeed present and thus likely relevant to model interpretation.
Although SW distance does pass the specificity test, we observed that, like the widely-cited CKA, it does not pass the sensitivity test of \citet{ding2021grounding}.
We expect this is in part due to the previously noted differences between the standard representation and our sampled point cloud; however, our sampling approach is needed to mitigate the computational costs of PH, which scale with dimensionality of the underlying space.

In future work, we plan to further characterize the types of topological structures present in hidden layers of CNNs, explore theoretical justifications for the success of our experiments, and complete a more thorough analysis of the sensitivity of the SW distance via principal component removal.
Finally, in order to aid DL practitioners in unlocking the hidden structures of their models, we plan to implement our methods into user-friendly tools.

\appendix
\section{Technical Appendix}
\subsection{Additional Figures for Mapper Experiments}
Here we include some additional figures that further strengthen our observations on foreground and background activation point clouds and their robustness to image noise.

\begin{figure*}[t]
    \centering
    \includegraphics[width=0.99\textwidth]{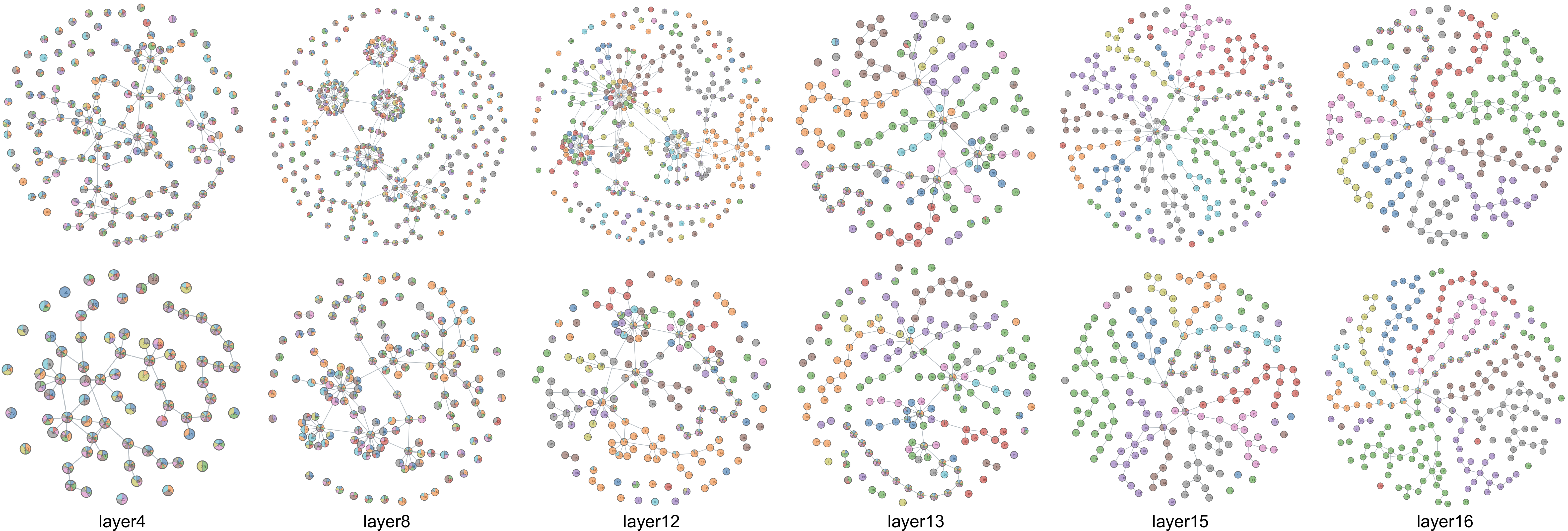}
    \caption{Mapper graphs generated from the foreground (top) and background (bottom) activations with top 5 largest weights.}
    \label{fig:mapper-fg-bg-top5}
\end{figure*}

In the main body of the paper we showed mapper graphs generated from the top 1 foreground and background activations.
In Figure \ref{fig:mapper-fg-bg-top5} we show additional mapper graphs from the top 5 foreground and background activations.
The branching properties and conclusions are similar to those for the top 1 foreground and background mapper graphs.

\begin{figure*}[ht]
    \centering
    \includegraphics[width=0.99\textwidth]{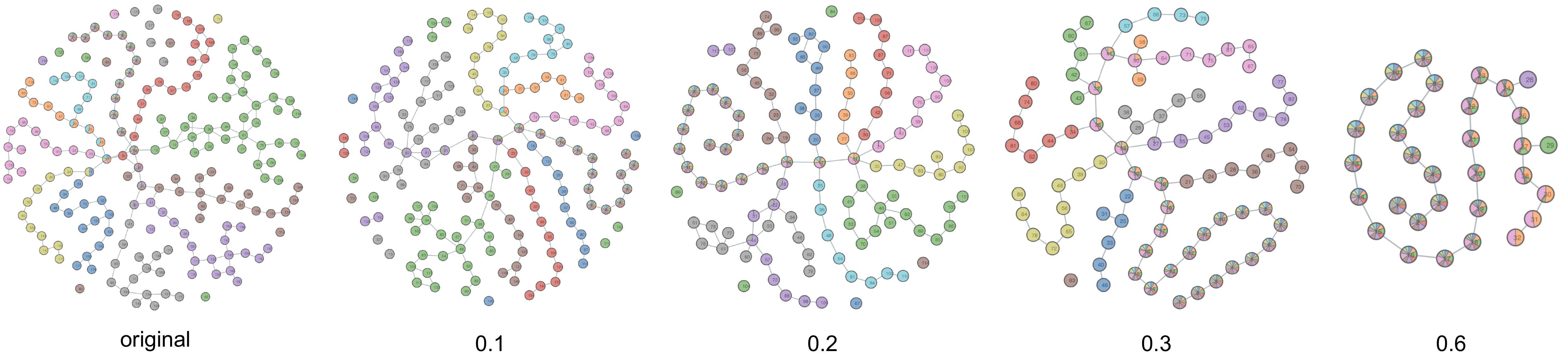}
    \caption{Perturbed mapper graphs generated from the top 5 foreground activations at the last convolutional layer.}
    \label{fig:mapper-noises-top5}
\end{figure*}

For our image perturbation experiment we showed mapper graphs generated from the full and foreground activations in the last layer for different levels of injected noise in the input images.
In Figure \ref{fig:mapper-noises-top5} we show additional mapper graphs for the top 5 full activations, again for only the last convolutional layer.
We observe that even at noise level $\sigma=0.3$ there is good branching structure in the last layer while the mapper graphs for the same noise level as in Figure \ref{fig:mapper-noises} have less clear branching. 
This may indicate that top 5 activations are more robust to input noise than full and foreground top 1 activations.

\begin{figure}[htbp]
    \centering
    \includegraphics[width=1.0\columnwidth]{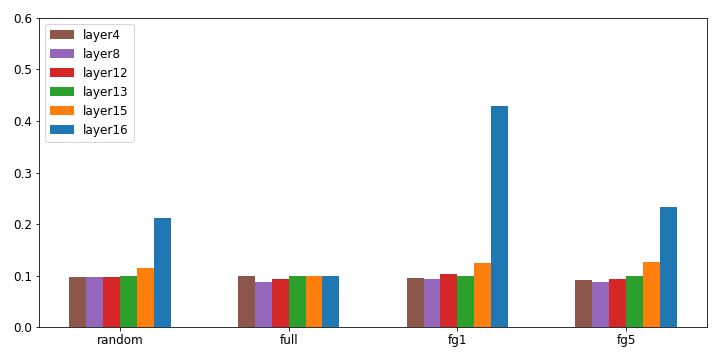}
    \caption{Class-wise purity of the \textbf{airplane} class for random, full, and foreground (top 1 and 5) at a variety of layers.}
    \label{fig:class-airplane-purity}
\end{figure}

\begin{figure}[htbp]
    \centering
    \includegraphics[width=1.0\columnwidth]{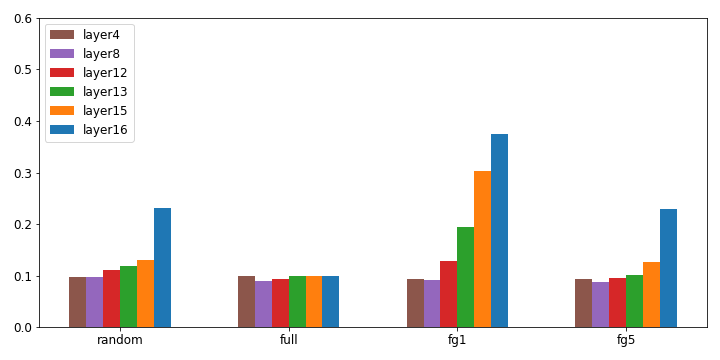}
    \caption{Class-wise purity of the \textbf{automobile} class for random, full, and foreground (top 1 and 5) at a variety of layers.}
    \label{fig:class-auto-purity}
\end{figure}

\begin{figure}[htbp]
    \centering
    \includegraphics[width=1.0\columnwidth]{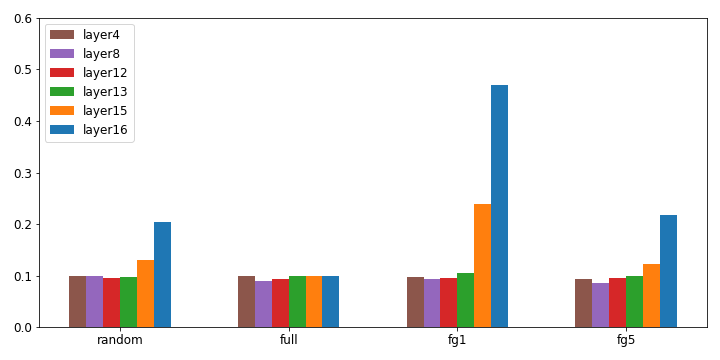}
    \caption{Class-wise purity of the \textbf{bird} class for random, full, and foreground (top 1 and 5) at a variety of layers.}
    \label{fig:class-bird-purity}
\end{figure}

\begin{figure}[htbp]
    \centering
    \includegraphics[width=1.0\columnwidth]{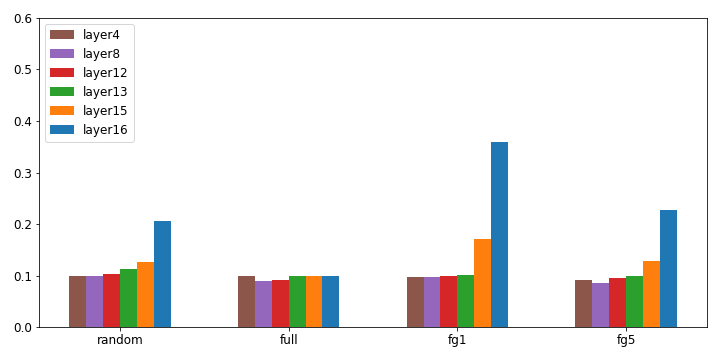}
    \caption{Class-wise purity of the \textbf{cat} class for random, full, and foreground (top 1 and 5) at a variety of layers.}
    \label{fig:class-cat-purity}
\end{figure}

\begin{figure}[htbp]
    \centering
    \includegraphics[width=1.0\columnwidth]{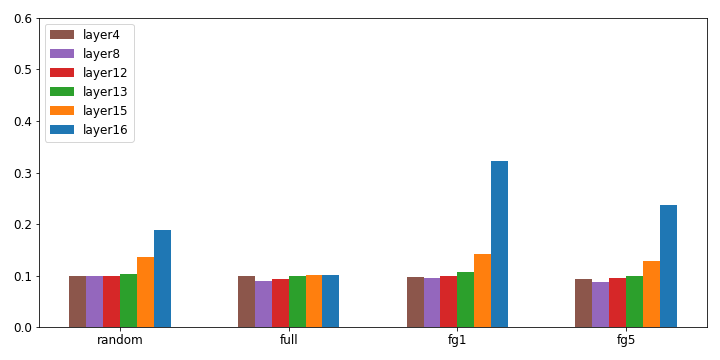}
    \caption{Class-wise purity of the \textbf{deer} class for random, full, and foreground (top 1 and 5) at a variety of layers.}
    \label{fig:class-deer-purity}
\end{figure}

\begin{figure}[htbp]
    \centering
    \includegraphics[width=1.0\columnwidth]{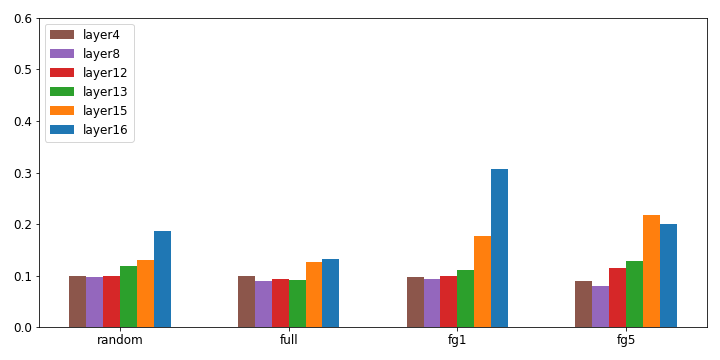}
    \caption{Class-wise purity of the \textbf{dog} class for random, full, and foreground (top 1 and 5) at a variety of layers.}
    \label{fig:class-dog-purity}
\end{figure}

\begin{figure}[htbp]
    \centering
    \includegraphics[width=1.0\columnwidth]{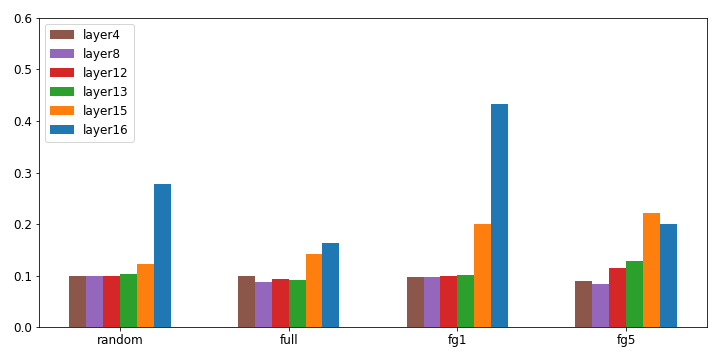}
    \caption{Class-wise purity of the \textbf{frog} class for random, full, and foreground (top 1 and 5) at a variety of layers.}
    \label{fig:class-frog-purity}
\end{figure}

\begin{figure}[ht]
    \centering
    \includegraphics[width=1.0\columnwidth]{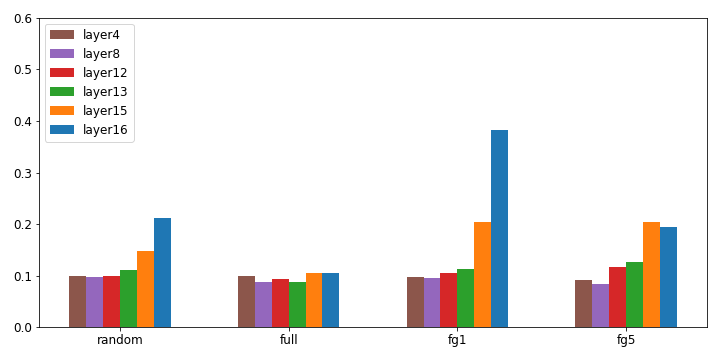}
    \caption{Class-wise purity of the \textbf{horse} class for random, full, and foreground (top 1 and 5) at a variety of layers.}
    \label{fig:class-horse-purity}
\end{figure}

\begin{figure}[ht]
    \centering
    \includegraphics[width=1.0\columnwidth]{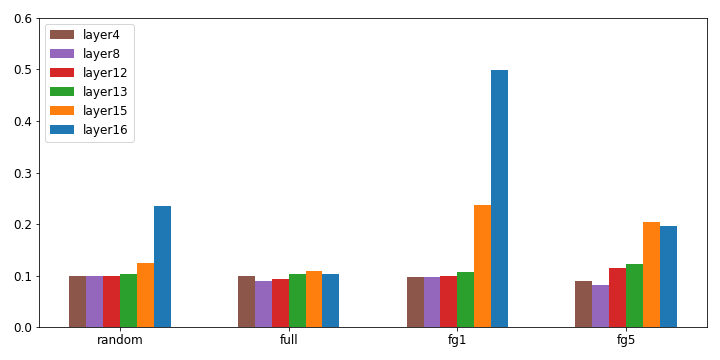}
    \caption{Class-wise purity of the \textbf{ship} class for random, full, and foreground (top 1 and 5) at a variety of layers.}
    \label{fig:class-ship-purity}
\end{figure}

\begin{figure}[ht]
    \centering
    \includegraphics[width=1.0\columnwidth]{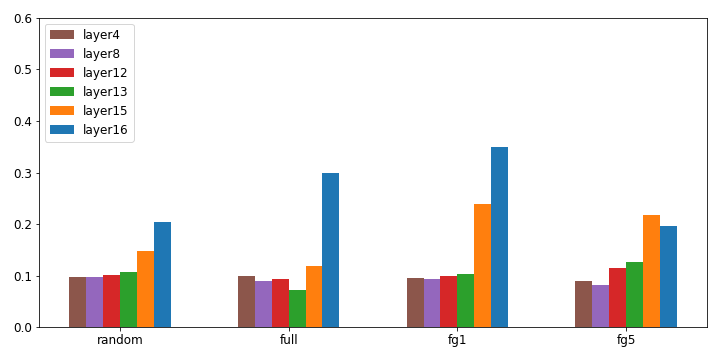}
    \caption{Class-wise purity of the \textbf{truck} class for random, full, and foreground (top 1 and 5) at a variety of layers.}
    \label{fig:class-truck-purity}
\end{figure}

Finally, we provide figures for the class-wise purity measures across all classes.
In the main body of the paper we showed class-wise purity for the deer class across a variety of layers and for random, full, and foreground (top 1 and top 5) spatial activation sampling.
Figures \ref{fig:class-airplane-purity} - \ref{fig:class-truck-purity} show the same class-wise purity plot for all 10 classes across all sampling methods with the same scale. The value range for the class-wise purity is $[0, 1]$. However, to avoid too much white space in the visualizations we make the scale in the plots to be $[0, 0.6]$, a bit more than the maximum purity observed.

\subsection{Reproducibility Checklist}
Many of the items in the full reproducibility checklist are addressed in the main body of the paper. Here we provide more details on model parameters, preprocessing and randomization code, and mapper graph parameters for the purposes of reproducibility of our results.

\paragraph{6.1:} Code for preprocessing the data is contained in the following supplementary code files: 
\begin{itemize}
    \item ResNet-18 models implemented and trained from scratch
    \begin{itemize}
        \item PH experiments:
        \begin{itemize}
            \item \texttt{cifar\_resnet.py}
            \item \texttt{ffcv\_cifar10\_train.py}
        \end{itemize}
        \item Mapper experiments: 
        \begin{itemize}
            \item \texttt{cifar\_train.py}
        \end{itemize}
    \end{itemize}

    \item Injecting Gaussian noise to input data: 
    \begin{itemize}
        \item \texttt{cifar\_\-extract\_\-full\_\-activations\_\-noises.py}
    \end{itemize}
    \item All types of point cloud creation for both PH and mapper experiments:
    \begin{itemize}
      \item Pulling top $l^2$-norm activation vectors from each layer for the Experiments with PH section:
      \begin{itemize}
          \item \texttt{topL2\_pointclouds.py}
      \end{itemize}
      \item Pulling activation vectors for the Experiments with Mapper section
      \begin{itemize}
      	\item Functions shared with all following scripts: \\
      	\texttt{cifar\_extract.py}
	\item Full activations: \\
	\texttt{cifar\_extract\_full\_activations.py}
	\item Random activations: \\
	\texttt{cifar\_extract\_sampled\_activations.py}
	\item Foreground (top 1, top 5) activations: \\
	\texttt{cifar\_\-extract\_\-full\_\-activations\_\-foreground.py}
	\item Background (top 1, top 5) activations:\\
	 \texttt{cifar\_\-extract\_\-full\_\-activations\_\-background.py}
	 \item Full, foreground activations with Gaussian noises:\\
	 \texttt{cifar\_\-extract\_\-full\_\-activations\_\-noises.py}\\
	 \texttt{cifar\_\-extract\_\-full\_\-activations\_\-foreground\_\-noises.py}
	 \item Activations of ImageNet from additional models:
	 \texttt{model-forward-pass.ipynb}
      \end{itemize}
    \end{itemize}
\end{itemize}

\paragraph{6.2:} Code for conducting and analyzing experiments uses the following custom scripts, openly available packages, and open source tools:
\begin{itemize}
  \item To compute PD given a point cloud we use the \texttt{ripser} Python package (function \texttt{ripser.ripser}) using default parameter choices \cite{ctralie2018ripser}
  
  \item To compute the SW distance we use the \texttt{persim} Python package (function \texttt{persim.sliced\_wasserstein}) using default parameter choices
  
  \item Code for Experiments with PH:
  \begin{itemize}
      \item SW distances between layers, for a single model or across differently initialized models for specificity test:
      \begin{itemize}
          \item \texttt{SW\_distances.py}
      \end{itemize}
      \item Removing principal components of point clouds for sensitivity test:
      \begin{itemize}
          \item \texttt{sensitivity\_testing.py}
      \end{itemize}
  \end{itemize}
  \item Computing mapper graphs
  \begin{itemize}
  	\item Mapper Interactive command line API:
	\begin{itemize}
		\item \texttt{cover.py}
		\item \texttt{kmapper.py}
		\item \texttt{mapper\_CLI.py}
		\item \texttt{nerve.py}
		\item \texttt{visuals.py}
	\end{itemize}
	\item The ``elbow'' approach to get the \texttt{eps} parameter value:
	\begin{itemize}
	    \item \texttt{get\_knn.py}
	\end{itemize}
	\item Full activations:
	\begin{itemize}
	    \item \texttt{get\_mapper\_full\_batches.py}
	\end{itemize}
	\item Random activations:
	\begin{itemize}
	    \item \texttt{get\_mapper\_random\_batches.py}
	\end{itemize}
	\item Foreground activations: 
	\begin{itemize}
		\item top 1: \texttt{get\_mapper\_full\_fg\_1.py}
		\item top 5: \texttt{get\_mapper\_full\_fg\_5.py}
	\end{itemize}
	\item Background activations: 
	\begin{itemize}
		\item top 1: \texttt{get\_mapper\_full\_bg\_1.py}
		\item top 5: \texttt{get\_mapper\_full\_bg\_5.py}
	\end{itemize}
	\item Full activations with Gaussian noises:
	\begin{itemize}
	    \item \texttt{get\_\-mapper\_\-full\_\-batches\_\-with\_\-noises.py}
	\end{itemize}
	\item Foreground activations with Gaussian noises: 
	\begin{itemize}
		\item top 1: \texttt{get\_mapper\_full\_fg\_1\_noises.py}
		\item top 5: \texttt{get\_mapper\_full\_fg\_5\_noises.py}
	\end{itemize}
	\item Activations from additional models of ImageNet:
	\begin{itemize}
	    \item \texttt{get\_mapper\_additional\_models.py}
	\end{itemize}
	
  \end{itemize}
  
  \item To compute the mapper graphs using MapperInteractive there are four important parameters to be tuned in the interactive interface: the number of intervals and the overlap rate to create the cover; the DBSCAN clustering parameters \texttt{eps} which sets the maximum distance between two points for one to be considered as in the neighborhood of the other; and \texttt{min\_samples}, the number of points in a neighborhood for a point to be considered as a core point.
  
  To create the mapper graphs we use MapperIneractive with the following parameter choices:
  \begin{itemize}
    \item \texttt{num\_intervals=40}
    \item \texttt{overlap\_rate=25\%}
    \item \texttt{min\_samples=5}
    \item For the mapper graphs generated from the random, full, foreground and background activations of the CIFAR-10 images at layers 16, 15, 13, 12, 8, and 4, the choices of \texttt{eps} are listed in Table~\ref{table:mapper-parameter-eps-cifar}
    \item For the mapper graphs generated from the perturbed images, the \texttt{eps} values are the same as those generated from the original images for comparison purpose.
    \item For the mapper graphs generated from the random and foreground activations of the ImageNet dataset at the last layers of models ResNet-18, Inception V1, Inception V3 and AlexNet, the choices of \texttt{eps} are listed in Table~\ref{table:mapper-parameter-eps-imagenet}
  \end{itemize}
  \item Computing purity measures from a mapper graph and a labeling of points
  \begin{itemize}
  	\item Node-wise purity: \texttt{get\_nodewise\_purity.py}
	\item Class-wise purity: \texttt{get\_classwise\_purity.py}
  \end{itemize}
\end{itemize}

\begin{table}[htbp]
    \centering
    \begin{tabular}{|c|c|c|c|c|c|c|}
    \hline
      \bf{Layer}& \bf{16} & \bf{15} & \bf{13} & \bf{12} & \bf{8} & \bf{4}\\ \hline
     random & 8.71  & 4.22 & 5.04 & 7.69 & 6.80 & 4.50\\ \hline
     full & 8.52 & 2.50 & 3.50 & 5.41 & 4.50 & 3.50 \\ \hline
     fg (top 1) & 10.65 & 8.50 & 9.29 & 11.87 & 8.51 & 4.99\\ \hline
     fg (top 5) & 11.00 & 7.50 & 9.52 & 10.05 & 7.00 & 4.00\\ \hline
     bg (top 1) & 12.09 & 8.19 & 9.20 & 12.41 & 8.20 & 4.85\\ \hline
     bg (top 5) & 10.07 & 8.50 & 9.55 & 11.02 & 7.57 & 4.52\\ \hline
    \end{tabular}
    \caption{The \texttt{eps} values for the mapper graphs generated from the random, full, top 1 and top 5 foreground and background activations of the CIFAR-10 images at layers 16, 15, 13, 12, 8 and 4.}
    \label{table:mapper-parameter-eps-cifar}
\end{table}

\begin{table}[htbp]
    \centering
    \begin{tabular}{|c|c|c|c|c|}
    \hline
      \bf{Model}& random & foreground \\ \hline
      \bf{ResNet-18} & 51.5 &  55.0  \\ \hline
     \bf{Inception V1} & 25.0 &  38.5 \\ \hline
      \bf{Inception V3} & 45.0 &  56.0  \\ \hline
       \bf{AlexNet} & 37.0 &  35.0  \\ \hline
    \end{tabular}
    \caption{The \texttt{eps} values for the mapper graphs generated from the random and foreground activations of the ImageNet at the last layer of the models ResNet-18, Inception V1, Inception V3 and AlexNet.}
    \label{table:mapper-parameter-eps-imagenet}
\end{table}

\paragraph{6.3:} All scripts and code outlined in this reproducibility checklist will be publicly released with a permissive license upon publication of this paper.

\paragraph{6.4:} The only code implementing new methods is for the purity measures which was already noted in reproducibility checklist item 6.2.

\paragraph{6.5:} Where randomness is employed in applying Gaussian noise we use \texttt{numpy.random.randint()} to generate random seeds. When selecting random batches of $N=1000$ test images in the PH section we use a PyTorch \texttt{DataLoader} with \texttt{shuffle=True} and a random seed of 0. All seeds are applied using \texttt{torch.manual\_seed()}.

\paragraph{6.6:} For our experiments we used the following computing infrastructures:
\begin{itemize}
    \item PH experiments:
    \begin{itemize}
        \item GPU models (training): NVIDIA DGX-A100 with 8 A100 GPUs
        \item GPU models (experiments): Dual NVIDIA P100 12GB PCI-e based GPU
        \item CPU models: 16 Dual Intel Broadwell E5-2620 v4 @ 2.10GHz CPUs
        \item Amount of memory: 64 GB 2133Mhz DDR4
        \item Operating system: Centos 7.8 based operating system (ROCKS 7)
        \item Relevant software libraries and frameworks:
        \begin{itemize}
            \item FFCV \cite{leclerc2022ffcv}
            \item PyTorch \cite{pytorch}
            \item Torchvision \cite{torchvision}
            \item Ripser \cite{ctralie2018ripser}
            \item Persim \cite{scikittda2019}
        \end{itemize}
    \end{itemize}
    \item Mapper experiments using ResNet-18 on CIFAR-10: 
    \begin{itemize}
    \item GPU models: NVIDIA 4x TITAN V with CUDA 11.2
    	\item CPU models: 32 Intel Xeon Silver 4108 CPU @ 1.80GHz cores (HT)
	    \item Amount of memory: 132GB of RAM
	    \item Operating system: OpenSUSE Leap 15.3 (x86\_64)
	    \item Relevant software libraries and frameworks: 
    	\begin{itemize}
    		\item Python (v3.6.15)
    		\item MapperInteractive
    		\item PyTorch (v1.9.0)
    		\item sklearn (v0.24.2)
    	\end{itemize}
    \end{itemize}
    \item Mapper experiments using ResNet-18, Inception V1, Inception V3 and AlexNet on ImageNet: 
    \begin{itemize}
    	\item GPU models: NVIDIA GTX 1060
	\item CPU models: Intel Xeon CPU E5-2630 v3 @ 2.40GHz
	    \item Amount of memory: 32GB of RAM
	    \item Operating system: OpenSUSE Leap 15.1
	    \item Relevant software libraries and frameworks: 
    	\begin{itemize}
    		\item Python (v3.7.5)
    		\item MapperInteractive
    		\item Pytorch (v1.4.0)
    		\item sklearn (v0.23.2)
    	\end{itemize}
    \end{itemize}
\end{itemize}

\paragraph{6.10:} Our comparison against other methods (e.g., SW distance vs. CCA and CKA) is not a head-to-head performance comparison and so comparison metrics are not applicable.
Instead we discuss the similarities and differences between our observations on SW distance and trends observed by \citet{ding2021grounding} on CCA and CKA.

\paragraph{6.11:} 
All architectures and hyper-parameters for the models are standard choices. For all models in the PH section we use the standard ResNet-18 architecture outlined by \citet{resnet} in their analysis of CIFAR-10. For the mapper experiments on the CIFAR-10 dataset, we trained a ResNet-18 model that we implement from scratch. 
For the mapper experiments on the ImageNet dataset, all the models (ResNet-18, Inception V1, Inception V3 and AlexNet) used are the pre-trained models from the PyTorch built-in model library without any modifications.

Parameters for creating mapper graphs were explained under item 6.2 above. 

To create our mapper graphs the final parameter choices were a result of the following process.
The \texttt{num\_intervals}, \texttt{overlap\_rate} and \texttt{min\_samples} were determined by manually tuning, and \texttt{eps} values were determined by sorting the distances of the $k$-th nearest neighbor for all points and finding the elbow point, where $k= \texttt{min\_samples}$.

\section*{Acknowledgements} 
MS, BW, and YZ are key contributors to this work. 
BW was partially funded by NSF DMS 2134223 and IIS 2205418.

\bibliography{refs_aaai23}

\begin{thebibliography}{36}
\providecommand{\natexlab}[1]{#1}

\bibitem[{Barannikov et~al.(2022)Barannikov, Trofimov, Balabin, and
  Burnaev}]{barranikov2022representation}
Barannikov, S.; Trofimov, I.; Balabin, N.; and Burnaev, E. 2022.
\newblock Representation Topology Divergence: A Method for Comparing Neural
  Network Representations.
\newblock In Chaudhuri, K.; Jegelka, S.; Song, L.; Szepesvari, C.; Niu, G.; and
  Sabato, S., eds., \emph{Proceedings of the 39th International Conference on
  Machine Learning}, volume 162 of \emph{Proceedings of Machine Learning
  Research}, 1607--1626. PMLR.

\bibitem[{Bradski(2000)}]{opencv_library}
Bradski, G. 2000.
\newblock {The OpenCV Library}.
\newblock \emph{Dr. Dobb's Journal of Software Tools}.

\bibitem[{Carri{\`e}re, Cuturi, and Oudot(2017)}]{pmlr-v70-carriere17a}
Carri{\`e}re, M.; Cuturi, M.; and Oudot, S. 2017.
\newblock Sliced {W}asserstein Kernel for Persistence Diagrams.
\newblock In Precup, D.; and Teh, Y.~W., eds., \emph{Proceedings of the 34th
  International Conference on Machine Learning}, volume~70 of \emph{Proceedings
  of Machine Learning Research}, 664--673. PMLR.

\bibitem[{Chen et~al.(2017)Chen, Papandreou, Schroff, and
  Adam}]{chen2017rethinking}
Chen, L.-C.; Papandreou, G.; Schroff, F.; and Adam, H. 2017.
\newblock Rethinking Atrous Convolution for Semantic Image Segmentation.
\newblock arXiv preprint arXiv:1706.05587.

\bibitem[{Deng et~al.(2009)Deng, Dong, Socher, Li, Li, and
  Fei-Fei}]{deng2009imagenet}
Deng, J.; Dong, W.; Socher, R.; Li, L.-J.; Li, K.; and Fei-Fei, L. 2009.
\newblock {ImageNet}: A large-scale hierarchical image database.
\newblock In \emph{2009 IEEE Conference on Computer Vision and Pattern
  Recognition}, 248--255.

\bibitem[{Ding, Denain, and Steinhardt(2021)}]{ding2021grounding}
Ding, F.; Denain, J.-S.; and Steinhardt, J. 2021.
\newblock Grounding Representation Similarity Through Statistical Testing.
\newblock In Ranzato, M.; Beygelzimer, A.; Dauphin, Y.; Liang, P.; and Vaughan,
  J.~W., eds., \emph{Advances in Neural Information Processing Systems},
  volume~34, 1556--1568.

\bibitem[{Edelsbrunner and Harer(2008)}]{edelsbrunner2008persistent}
Edelsbrunner, H.; and Harer, J. 2008.
\newblock Persistent homology---a survey.
\newblock In \emph{Surveys on Discrete and Computational Geometry}, volume 453,
  257--282. American Mathematical Society.

\bibitem[{Gabrielsson and Carlsson(2019)}]{gabrielsson2019exposition}
Gabrielsson, R.~B.; and Carlsson, G. 2019.
\newblock Exposition and Interpretation of the Topology of Neural Networks.
\newblock In \emph{2019 18th IEEE International Conference on Machine Learning
  and Applications (ICMLA)}, 1069--1076.

\bibitem[{Gebhart, Schrater, and Hylton(2019)}]{gebhart2019characterizing}
Gebhart, T.; Schrater, P.; and Hylton, A. 2019.
\newblock Characterizing the Shape of Activation Space in Deep Neural Networks.
\newblock \emph{2019 18th IEEE International Conference On Machine Learning And
  Applications (ICMLA)}, 1537--1542.

\bibitem[{Ghrist(2008)}]{ghrist2008barcodes}
Ghrist, R. 2008.
\newblock Barcodes: the persistent topology of data.
\newblock \emph{Bulletin of the American Mathematical Society (New Series)},
  45(1): 61--75.

\bibitem[{Guss and Salakhutdinov(2018)}]{guss2018characterizing}
Guss, W.~H.; and Salakhutdinov, R. 2018.
\newblock On Characterizing the Capacity of Neural Networks using Algebraic
  Topology.
\newblock arXiv preprint arXiv:1802.04443.

\bibitem[{He et~al.(2016)He, Zhang, Ren, and Sun}]{resnet}
He, K.; Zhang, X.; Ren, S.; and Sun, J. 2016.
\newblock Deep Residual Learning for Image Recognition.
\newblock In \emph{Proceedings of the IEEE Conference on Computer Vision and
  Pattern Recognition (CVPR)}.

\bibitem[{Kim et~al.(2018)Kim, Wattenberg, Gilmer, Cai, Wexler, Viegas, and
  sayres}]{kim2018interpretability}
Kim, B.; Wattenberg, M.; Gilmer, J.; Cai, C.; Wexler, J.; Viegas, F.; and
  sayres, R. 2018.
\newblock Interpretability Beyond Feature Attribution: Quantitative Testing
  with Concept Activation Vectors ({TCAV}).
\newblock In Dy, J.; and Krause, A., eds., \emph{Proceedings of the 35th
  International Conference on Machine Learning}, volume~80 of \emph{Proceedings
  of Machine Learning Research}, 2668--2677. PMLR.

\bibitem[{Kornblith et~al.(2019)Kornblith, Norouzi, Lee, and
  Hinton}]{kornblith2019similarity}
Kornblith, S.; Norouzi, M.; Lee, H.; and Hinton, G. 2019.
\newblock Similarity of Neural Network Representations Revisited.
\newblock In Chaudhuri, K.; and Salakhutdinov, R., eds., \emph{Proceedings of
  the 36th International Conference on Machine Learning}, volume~97 of
  \emph{Proceedings of Machine Learning Research}, 3519--3529. PMLR.

\bibitem[{Krizhevsky and Hinton(2009)}]{cifar10}
Krizhevsky, A.; and Hinton, G. 2009.
\newblock Learning multiple layers of features from tiny images.
\newblock Technical Report TR-2009, University of Toronto, Toronto, Ontario.

\bibitem[{Krizhevsky, Sutskever, and Hinton(2012)}]{krizhevsky2012imagenet}
Krizhevsky, A.; Sutskever, I.; and Hinton, G.~E. 2012.
\newblock ImageNet Classification with Deep Convolutional Neural Networks.
\newblock In Pereira, F.; Burges, C.; Bottou, L.; and Weinberger, K., eds.,
  \emph{Advances in Neural Information Processing Systems}, volume~25.

\bibitem[{Lacombe, Ike, and Umeda(2021)}]{lacombe2021topological}
Lacombe, T.; Ike, Y.; and Umeda, Y. 2021.
\newblock Topological Uncertainty: Monitoring trained neural networks through
  persistence of activation graphs.
\newblock In \emph{Proceedings of the 30th International Joint Conference on
  Artificial Intelligence}, 2666--2672.

\bibitem[{Leclerc et~al.(2022)Leclerc, Ilyas, Engstrom, Park, Salman, and
  Madry}]{leclerc2022ffcv}
Leclerc, G.; Ilyas, A.; Engstrom, L.; Park, S.~M.; Salman, H.; and Madry, A.
  2022.
\newblock {FFCV}: Accelerating Training by Removing Data Bottlenecks.
\newblock \url{https://github.com/libffcv/ffcv/}.
\newblock Commit f253865.

\bibitem[{Mahendran and Vedaldi(2015)}]{mahendran2015understanding}
Mahendran, A.; and Vedaldi, A. 2015.
\newblock Understanding Deep Image Representations by Inverting Them.
\newblock In \emph{Proceedings of the IEEE Conference on Computer Vision and
  Pattern Recognition (CVPR)}.

\bibitem[{Marcel and Rodriguez(2010)}]{torchvision}
Marcel, S.; and Rodriguez, Y. 2010.
\newblock Torchvision the Machine-Vision Package of Torch.
\newblock In \emph{Proceedings of the 18th ACM International Conference on
  Multimedia}, MM~'10, 1485--1488. New York, NY, USA: Association for Computing
  Machinery.

\bibitem[{Morcos, Raghu, and Bengio(2018)}]{morcos2018insights}
Morcos, A.; Raghu, M.; and Bengio, S. 2018.
\newblock Insights on representational similarity in neural networks with
  canonical correlation.
\newblock In Bengio, S.; Wallach, H.; Larochelle, H.; Grauman, K.;
  Cesa-Bianchi, N.; and Garnett, R., eds., \emph{Advances in Neural Information
  Processing Systems}, volume~31.

\bibitem[{Olah et~al.(2020)Olah, Cammarata, Schubert, Goh, Petrov, and
  Carter}]{olah2020zoom}
Olah, C.; Cammarata, N.; Schubert, L.; Goh, G.; Petrov, M.; and Carter, S.
  2020.
\newblock Zoom In: An Introduction to Circuits.
\newblock \emph{Distill}.
\newblock Https://distill.pub/2020/circuits/zoom-in.

\bibitem[{Paszke et~al.(2019)Paszke, Gross, Massa, Lerer, Bradbury, Chanan,
  Killeen, Lin, Gimelshein, Antiga, Desmaison, Kopf, Yang, DeVito, Raison,
  Tejani, Chilamkurthy, Steiner, Fang, Bai, and Chintala}]{pytorch}
Paszke, A.; Gross, S.; Massa, F.; Lerer, A.; Bradbury, J.; Chanan, G.; Killeen,
  T.; Lin, Z.; Gimelshein, N.; Antiga, L.; Desmaison, A.; Kopf, A.; Yang, E.;
  DeVito, Z.; Raison, M.; Tejani, A.; Chilamkurthy, S.; Steiner, B.; Fang, L.;
  Bai, J.; and Chintala, S. 2019.
\newblock PyTorch: An Imperative Style, High-Performance Deep Learning Library.
\newblock In Wallach, H.; Larochelle, H.; Beygelzimer, A.; dAlch\'{e} Buc, F.;
  Fox, E.; and Garnett, R., eds., \emph{Advances in Neural Information
  Processing Systems 32}, 8024--8035. Curran Associates, Inc.

\bibitem[{Raghu et~al.(2017)Raghu, Gilmer, Yosinski, and
  Sohl-Dickstein}]{raghu2017svcca}
Raghu, M.; Gilmer, J.; Yosinski, J.; and Sohl-Dickstein, J. 2017.
\newblock {SVCCA: Singular Vector Canonical Correlation Analysis for Deep
  Learning Dynamics and Interpretability}.
\newblock In Guyon, I.; Luxburg, U.~V.; Bengio, S.; Wallach, H.; Fergus, R.;
  Vishwanathan, S.; and Garnett, R., eds., \emph{Advances in Neural Information
  Processing Systems}, volume~30.

\bibitem[{Rathore et~al.(2021)Rathore, Chalapathi, Palande, and
  Wang}]{rathore2021topoact}
Rathore, A.; Chalapathi, N.; Palande, S.; and Wang, B. 2021.
\newblock TopoAct: Visually Exploring the Shape of Activations in Deep
  Learning.
\newblock \emph{Computer Graphics Forum}, 40(1): 382--397.

\bibitem[{Rieck et~al.(2019)Rieck, Togninalli, Bock, Moor, Horn, Gumbsch, and
  Borgwardt}]{rieck2019neural}
Rieck, B.; Togninalli, M.; Bock, C.; Moor, M.; Horn, M.; Gumbsch, T.; and
  Borgwardt, K. 2019.
\newblock Neural Persistence: {A} Complexity Measure for Deep Neural Networks
  Using Algebraic Topology.
\newblock In \emph{International Conference on Learning
  Representations~(ICLR)}.

\bibitem[{Saul and Tralie(2019)}]{scikittda2019}
Saul, N.; and Tralie, C. 2019.
\newblock {Scikit-TDA}: Topological Data Analysis for Python.

\bibitem[{Selvaraju et~al.(2017)Selvaraju, Cogswell, Das, Vedantam, Parikh, and
  Batra}]{selvaraju2017grad}
Selvaraju, R.~R.; Cogswell, M.; Das, A.; Vedantam, R.; Parikh, D.; and Batra,
  D. 2017.
\newblock {Grad-CAM}: Visual Explanations From Deep Networks via Gradient-Based
  Localization.
\newblock In \emph{Proceedings of the IEEE International Conference on Computer
  Vision (ICCV)}.

\bibitem[{Singh, Memoli, and Carlsson(2007)}]{singh2007topological}
Singh, G.; Memoli, F.; and Carlsson, G. 2007.
\newblock {Topological Methods for the Analysis of High Dimensional Data Sets
  and 3D Object Recognition}.
\newblock In Botsch, M.; Pajarola, R.; Chen, B.; and Zwicker, M., eds.,
  \emph{Eurographics Symposium on Point-Based Graphics}.

\bibitem[{Szegedy et~al.(2015)Szegedy, Liu, Jia, Sermanet, Reed, Anguelov,
  Erhan, Vanhoucke, and Rabinovich}]{szegedy2015going}
Szegedy, C.; Liu, W.; Jia, Y.; Sermanet, P.; Reed, S.; Anguelov, D.; Erhan, D.;
  Vanhoucke, V.; and Rabinovich, A. 2015.
\newblock Going deeper with convolutions.
\newblock In \emph{2015 IEEE Conference on Computer Vision and Pattern
  Recognition (CVPR)}, 1--9.

\bibitem[{Szegedy et~al.(2016)Szegedy, Vanhoucke, Ioffe, Shlens, and
  Wojna}]{szegedy2016rethinking}
Szegedy, C.; Vanhoucke, V.; Ioffe, S.; Shlens, J.; and Wojna, Z. 2016.
\newblock Rethinking the Inception Architecture for Computer Vision.
\newblock In \emph{2016 IEEE Conference on Computer Vision and Pattern
  Recognition (CVPR)}, 2818--2826.

\bibitem[{Tralie, Saul, and Bar-On(2018)}]{ctralie2018ripser}
Tralie, C.; Saul, N.; and Bar-On, R. 2018.
\newblock {Ripser.py}: A Lean Persistent Homology Library for Python.
\newblock \emph{The Journal of Open Source Software}, 3(29): 925.

\bibitem[{Wei et~al.(2015)Wei, Zhou, Torrabla, and
  Freeman}]{wei2015understanding}
Wei, D.; Zhou, B.; Torrabla, A.; and Freeman, W. 2015.
\newblock Understanding intra-class knowledge inside CNN.
\newblock arXiv preprint arXiv:1507.02379.

\bibitem[{Wheeler, Bouza, and Bubenik(2021)}]{wheeler2021activation}
Wheeler, M.; Bouza, J.; and Bubenik, P. 2021.
\newblock Activation Landscapes as a Topological Summary of Neural Network
  Performance.
\newblock In \emph{2021 IEEE International Conference on Big Data (Big Data)},
  3865--3870. IEEE.

\bibitem[{Zhou et~al.(2015)Zhou, Khosla, Lapedriza, Oliva, and
  Torralba}]{zhou2015object}
Zhou, B.; Khosla, A.; Lapedriza, {\`{A}}.; Oliva, A.; and Torralba, A. 2015.
\newblock Object Detectors Emerge in Deep Scene CNNs.
\newblock In Bengio, Y.; and LeCun, Y., eds., \emph{3rd International
  Conference on Learning Representations, {ICLR} 2015, San Diego, CA, USA, May
  7-9, 2015, Conference Track Proceedings}.

\bibitem[{Zhou et~al.(2021)Zhou, Chalapathi, Rathore, Zhao, and
  Wang}]{zhou2021mapper}
Zhou, Y.; Chalapathi, N.; Rathore, A.; Zhao, Y.; and Wang, B. 2021.
\newblock Mapper Interactive: A Scalable, Extendable, and Interactive Toolbox
  for the Visual Exploration of High-Dimensional Data.
\newblock In \emph{2021 IEEE 14th Pacific Visualization Symposium
  (PacificVis)}, 101--110.

\end{thebibliography}

\end{document}